\PassOptionsToPackage{table}{xcolor}
\documentclass[letterpaper]{article} % DO NOT CHANGE THIS
\usepackage{aaai2026}  % DO NOT CHANGE THIS Add [submission] to disable anonymous
\usepackage{times}  % DO NOT CHANGE THIS
\usepackage{helvet}  % DO NOT CHANGE THIS
\usepackage{courier}  % DO NOT CHANGE THIS
\usepackage[hyphens]{url}  % DO NOT CHANGE THIS
\usepackage{graphicx} % DO NOT CHANGE THIS
\usepackage{natbib}  % DO NOT CHANGE THIS AND DO NOT ADD ANY OPTIONS TO IT
\usepackage{caption} % DO NOT CHANGE THIS AND DO NOT ADD ANY OPTIONS TO IT
\usepackage{algorithm}
\usepackage{algorithmic}

\usepackage{newfloat}
\usepackage{listings}
\DeclareCaptionStyle{ruled}{labelfont=normalfont,labelsep=colon,strut=off} % DO NOT CHANGE THIS
\floatstyle{ruled}
\newfloat{listing}{tb}{lst}{}
\floatname{listing}{Listing}
\usepackage{lipsum}

\usepackage[acronym]{glossaries}
\makeglossaries
\newacronym{cvrp}{CVRP}{Capacitated Vehicle Routing Problem}
\newacronym{tsp}{TSP}{Travelling Salesman Problem}
\newacronym{bpp}{BPP}{Bin Packing Problem}
\newacronym{mkp}{MKP}{Multiple Knapsack Problem}
\newacronym{op}{OP}{Orienteering Problem}
\newacronym{cop}{COP}{Combinatorial Optimization Problem}

\newacronym{gphh}{GPHH}{Genetic Programming Hyper Heuristic}
\newacronym{nco}{NCO}{Neural Combinatorial Optimization}
\newacronym{llm}{LLM}{Large Language Model}
\newacronym{reevo}{ReEvo}{Reflective Evolution}

\newacronym{ahd}{AHD}{Automatic Heuristic Design}
\newacronym{mcts}{MCTS}{Monte Carlo Tree Search}

\newacronym{mtcs-ahd}{MTCS-AHD}{Monte Carlo Tree Search-Automatic Heuristic Design}
\newacronym{lhh}{LHH}{Language Hyper-Heuristics}
\newacronym{cmcts}{CMCTS}{Competitive Monte Carlo Tree Search}
\newacronym{ucb}{UCB}{Upper Confidence Bound}
\newacronym{motif}{MOTIF}{\textbf{\underline{M}}ulti-strategy \textbf{\underline{O}}ptimization via \textbf{\underline{T}}urn-based \textbf{\underline{I}}nteractive \textbf{\underline{F}}ramework}

\newacronym{gls}{GLS}{Guided Local Search}
\newacronym{aco}{ACO}{Ant Colony Optimization}

\usepackage{svg}
\usepackage{booktabs} 
\usepackage{multirow}
\usepackage{makecell}
\usepackage{pifont}
\usepackage[most]{tcolorbox}
\usepackage{pythonhighlight}

\usepackage{listings}
\usepackage{amsthm}
\usepackage{paralist}
\usepackage{enumitem}
\theoremstyle{definition}
\newtheorem{definition}{Definition}[section]

\usepackage{amssymb}
\newtcolorbox{mybox1}[2][]{%
  colback=#1!5, colframe=#1!80, coltitle=white, colupper=black,
  fonttitle=\sffamily, fontupper=\sffamily,
  title=#2,
  arc=2mm, top=2mm, bottom=2mm, 
  boxrule=1pt, sharp corners,
  before skip=1em, after skip=1em,
  enhanced
}

\title{MOTIF: Multi-strategy Optimization via Turn-based Interactive Framework}

\author {
    Nguyen Viet Tuan Kiet\textsuperscript{\rm 1},
    Tung Dao\textsuperscript{\rm 1},
    Cong Dao Tran\textsuperscript{\rm 2},
    Huynh Thi Thanh Binh\textsuperscript{\rm 1}\footnote{Corresponding author}
}

\affiliations {
\textsuperscript{\rm 1}Hanoi University of Science and Technology, Vietnam\\
\textsuperscript{\rm 2}FPT Software AI Center, Vietnam\\

\{kiet.nvt220032, tung.dv242050M\}@sis.hust.edu.vn, 
daotc2@fpt.com, binhht@soict.hust.edu.vn}

\begin{document}

\maketitle

\begin{abstract}
Designing effective algorithmic components remains a fundamental obstacle in tackling NP-hard combinatorial optimization problems (COPs), where solvers often rely on carefully hand-crafted strategies. Despite recent advances in using large language models (LLMs) to synthesize high-quality components, most approaches restrict the search to a single element—commonly a heuristic scoring function—thus missing broader opportunities for innovation. In this paper, we introduce a broader formulation of solver design as a multi-strategy optimization problem, which seeks to jointly improve a set of interdependent components under a unified objective. To address this, we propose \textbf{M}ulti-strategy \textbf{O}ptimization via \textbf{T}urn-based \textbf{I}nteractive \textbf{F}ramework (\textbf{MOTIF})—a novel framework based on Monte Carlo Tree Search that facilitates turn-based optimization between two LLM agents. At each turn, an agent improves one component by leveraging the history of both its own and its opponent’s prior updates, promoting both competitive pressure and emergent cooperation. This structured interaction broadens the search landscape and encourages the discovery of diverse, high-performing solutions. Experiments across multiple COP domains show that MOTIF consistently outperforms state-of-the-art methods, highlighting the promise of turn-based, multi-agent prompting for fully automated solver design.
\end{abstract}

\begin{figure}[t]
    \centering
    % \includesvg[width=\columnwidth]{figures/Fig1}
    \includegraphics[width=0.9\columnwidth]{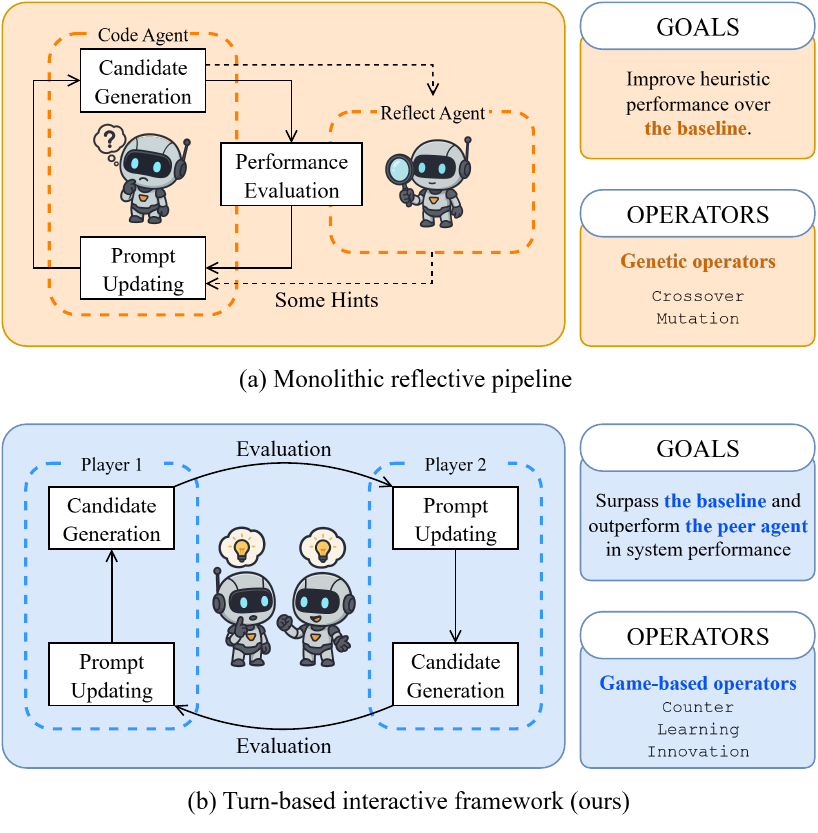}
    \caption{(a) \textit{Monolithic reflective pipeline}: generation and reflection exchange one‑way hints around a single evaluator with minimal behavioral awareness. (b) \textit{Turn‑based interactive framework}: two agents take turns generating and updating under shared evaluations, yielding explicit peer feedback, richer diversity, and adaptive explore–exploit balance.}
    \label{fig:first}
\end{figure}

\section{Introduction}

From vehicle routing and scheduling to circuit design and logistics, \gls{cop} underpin many of society’s most complex decision systems \cite{phan2025pareto, rajendran1993heuristic, tan2021comprehensive}. Yet, despite decades of progress, building effective solvers for \gls{cop}s remains a costly and highly manual endeavor—requiring domain-specific heuristics, iterative tuning, and substantial expert effort. Heuristics and meta-heuristics are currently the most common methods used to tackle these problems \cite{desale2015heuristic}. However, these techniques often require substantial expert input, which increases development costs and limits their flexibility across various problems.

Significant progress has been made in the field of \gls{ahd} \cite{burke2018classification}. A prominent example is \gls{gphh} \cite{langdon2013foundations}, a widely recognized algorithm for \gls{ahd}. However, many existing approaches still rely heavily on predefined heuristic spaces or rules designed by human experts \cite{pillay2018hyper}. Recently, \gls{lhh} has emerged as a promising solution to overcome this limitation. Rather than manually designing heuristic functions or relying on fixed search spaces, researchers have begun leveraging the reasoning capabilities of \gls{llm} to automatically generate more innovative heuristics. For instance, Liu et al. \cite{liu2024evolution, romera2024mathematical} integrated evolutionary algorithms with \gls{llm} to evolve code with minimal human intervention. Ye et al. introduced \gls{reevo} \cite{ye2024reevo}, a framework that combines evolutionary search with self-reflection \cite{shinn2023reflexion} to further improve the effectiveness of \gls{llm}. Subsequent studies, such as HSEvo \cite{dat2025hsevo}, incorporated diversity measurements and harmony search \cite{shi2012hybrid} to encourage more creative solutions. In addition to evolutionary methods, other search algorithms have also been explored in \gls{lhh}. For example, MCTS-AHD \cite{zheng2025monte} employed \gls{mcts} with progressive widening \cite{browne2012survey, coulom2007computing} to better explore the heuristic space. These methods, which require minimal expert knowledge, have achieved competitive performance compared to state-of-the-art solvers and neural network–based approaches.

% However, population-based methods still face the issue of premature convergence, often becoming stuck at local optima \cite{zheng2025monte}. This limitation prevents them from achieving near-optimal solutions for larger instances. To address this challenge, Zhang et al. proposed \gls{mtcs-ahd}, a tree-structured search over the heuristic space using \gls{llm}. This method improves exploration of the heuristic space by temporarily evolving underperforming heuristics while focusing on better-performing ones.

However, previous approaches have generally focused on searching for a single heuristic function within a general framework, which limits the creativity and exploration potential of \gls{llm}. To address this limitation, we propose a multi-strategy optimization approach that aims to optimize a set of strategies simultaneously rather than focusing on just one. This broader search space enables \gls{llm} to generate more diverse and innovative solutions. Moreover, because multi-strategy optimization requires a significantly more complex search mechanism than existing methods, we introduce a turn-based interactive framework in which two players sequentially compete to outperform each other in discovering the best set of strategies, as presented  in Figure~\ref{fig:first}. In conclusion, our contributions can be summarized as follows:
\begin{itemize}
    \item We introduce the novel problem of \textit{multi-strategy optimization}, where the goal is to jointly optimize a system of algorithmic components under a shared performance objective. This setting generalizes prior work that only targets single-component improvements.
    \item We propose a two-round competitive framework that decomposes the problem into (i) \textit{component-wise optimization} using \gls{cmcts} under a dynamic baseline, and (ii) \textit{system-aware refinement} in which agents compete in a turn-based manner with fixed baselines and full system visibility.
    \item We design a library of competitive operators—including \textit{counter}, \textit{learning}, and \textit{innovation}—that structure LLM prompting based on the opponent's behavior, recent history, and baseline context. This enables emergent self-play dynamics and yields superior implementations through adversarial pressure and contextual learning.
    \item We conduct extensive experiments across three algorithmic frameworks, five \gls{cop} domains, and diverse strategy sets. Results consistently demonstrate that MOTIF outperforms prior frameworks—both in single-strategy and multi-strategy optimization—validating the power of turn-based self-play.
\end{itemize}

\section{Related Works}
\label{gen_inst}
\subsection{Automatic Heuristics Design}
Previous work on \gls{lhh} \cite{liu2024evolution, ye2024reevo, dat2025hsevo, zheng2025monte, tranlarge} has explored the use of \gls{llm}s to design heuristic functions via evolutionary algorithms or tree-based search. While these approaches improve over traditional metaheuristics and \gls{nco}, they exhibit two core limitations. First, they aim to optimize a single heuristic in isolation, overlooking the potential of weaker components that, when combined, can yield stronger overall performance. This narrow focus restricts the generative flexibility of \gls{llm}s. Second, although prior frameworks incorporate self-reflection to assess interactions between heuristics, such mechanisms are limited in scope and depth. Typically based on static comparisons or surface-level summaries, they struggle to capture hidden incompatibilities or potential synergies. As a result, the LLM receives insufficient feedback for meaningful revisions, hampering its ability to conduct deeper exploratory optimization and refine its outputs throughout the search process.
\subsection{Learning Through Self-play}

Recent work has shown that game-based self-play can enhance LLM reasoning through structured interaction, such as critique, revision, or deception. For instance, SPAG \cite{cheng2025selfplayingadversariallanguagegame} frames reasoning as an attacker–defender game to improve deception detection; CDG \cite{wangimproving} trains a prover–critic pair to expose flawed reasoning; and \cite{fu2023improving} uses self-play and AI feedback in negotiation tasks to refine decision-making. While effective for internal rationality and dialogue skills, these methods do not address the co-evolution of multiple algorithmic components for structured tasks. In contrast, our approach applies self-play to multi-strategy optimization, where LLM agents iteratively critique, improve, and compete over distinct heuristics—introducing a new paradigm of inter-strategy competition under system-level feedback. To our knowledge, this is the first use of self-play for optimizing combinatorial solvers in this setting.

\section{Multi-strategy Optimization}
\label{headings}

Prior works \cite{liu2024evolution, ye2024reevo, dat2025hsevo, zheng2025monte} has largely focused on tuning a single heuristic assuming it alone drives solver quality. In contrast, we adopt a multi-strategy view that treats each routine as an independent optimization target, enabling coordinated improvements across the solver pipeline. This modular perspective fosters richer interaction among routines and often yields greater overall solver synergy than isolated rule refinement.

\begin{definition}[Domain, Instance, and Solution]
A \gls{cop} domain \(d\) defines a class of discrete problems, such as the \gls{tsp} or \gls{cvrp}. 
Without loss of generality, we assume the objective is to minimize a cost function; maximization problems can be equivalently transformed by negating the objective. Under this convention, each domain specifies:
\begin{itemize}
    \item a space of instances \(\mathcal{X}_d\), where each instance \(\mathbf{x} \in \mathcal{X}_d\) encodes problem-specific data (e.g., graph structure, distances matrix, constraints),
    \item a global solution space \(\mathcal{Y}_d\), representing all possible solutions across all instances, and
    \item an objective function \(f_d: \mathcal{X}_d \times \mathcal{Y}_d \rightarrow \mathbb{R}\) that quantifies the quality of a solution relative to a given instance.
\end{itemize}
For a specific instance \(\mathbf{x}   \), the feasible solutions form a subset \(\mathcal{Y}_d(\mathbf{x}) \subset \mathcal{Y}_d\).
The goal is to find a solution \(\mathbf{y}^* \in \mathcal{Y}_d(\mathbf{x})\) that minimizes \(f_d(\mathbf{x}, \mathbf{y})\).
\end{definition}

\begin{definition}[Solver and Strategy]
A solver \(s\) is an algorithmic framework that generates a solution \(\mathbf{y} \in \mathcal{Y}_d(\mathbf{x})\) for a given instance \(\mathbf{x} \in \mathcal{X}_d\), guided by a collection of internal routines. Each such routine is referred to as a strategy \(\pi_k\), which may include heuristic scoring rules, construction policies, neighborhood moves, penalty update mechanisms, or other algorithmic components. The solver operates as:
\begin{equation}
  \mathbf{y} = s\big(\,\mathbf{x} \mid (\pi_1,\pi_2,\ldots,\pi_K)\,\big).
\end{equation}
\end{definition}
\vspace{6 pt}
\begin{definition}[Strategy Space]
For each strategy type \(\pi\), let \(\mathcal{S}_\pi\) denote the space of all valid implementations, including both functional variants and parametrizations. 
All strategies in \(\mathcal{S}_\pi\) share a common function signature and optimization objective, ensuring interoperability within the solver. Variations among strategies arise from differences in internal logic, parameter choices, or heuristic design, while maintaining consistent input-output behavior.
\end{definition}

Given a solver \(s\) equipped with a sequence of $K$ strategies \(\mathbf{\Pi}=(\pi_1,\pi_2,\ldots,\pi_K)\), its performance on an instance \(\mathbf{x} \in \mathcal{X}_d\) is evaluated by the objective value 
\begin{equation}
F_d\bigr(\,\mathbf{x}\mid\mathbf{\Pi}\,\bigl)\,=f_d\Bigr(\,\mathbf{x}, s\bigr(\,\mathbf{x} \mid \mathbf{\Pi}\,\bigl)\Bigl).
\end{equation}
Since the solver's behavior is determined by its underlying strategies, optimizing solver quality amounts to optimizing the design of these strategies.

\begin{definition}[Multi-strategy Optimization]
\label{def:mo}
The multi-strategy optimization problem aims to simultaneously optimize a sequence of \(K\) strategies \(\mathbf{\Pi}=(\pi_1,\pi_2,\ldots,\pi_K)\), implemented within solver \(s\), where each \(\pi_k\) belongs to its corresponding strategy space \(\mathcal{S}_{\pi_k}\), to minimize the expected solver objective across the domain, as follows:
\begin{equation}
  \mathbf{\Pi}^*=\arg\min_{\mathbf{\Pi}} \;\; \mathbb{E}_{\mathbf{x} \sim \mathcal{X}_d} \Bigr[F_d\bigr(\,\mathbf{x}\mid\mathbf{\Pi}\,\bigl)\Bigl],
\end{equation}
subject to a total computational budget \(T\) (e.g., total solver runs, code evaluation time).
\end{definition}

Ours formulation enables a more holistic view of solver design: instead of optimizing components in isolation, we jointly evolve multiple, interacting strategies under a unified optimization objective. By modeling each strategy as an adaptive and parameterizable unit, our approach supports richer design spaces and unlocks synergies that static, hand-crafted routines cannot capture.

\section{MOTIF}

\begin{figure*}[t]
    \centering
    \includegraphics[width=\linewidth]{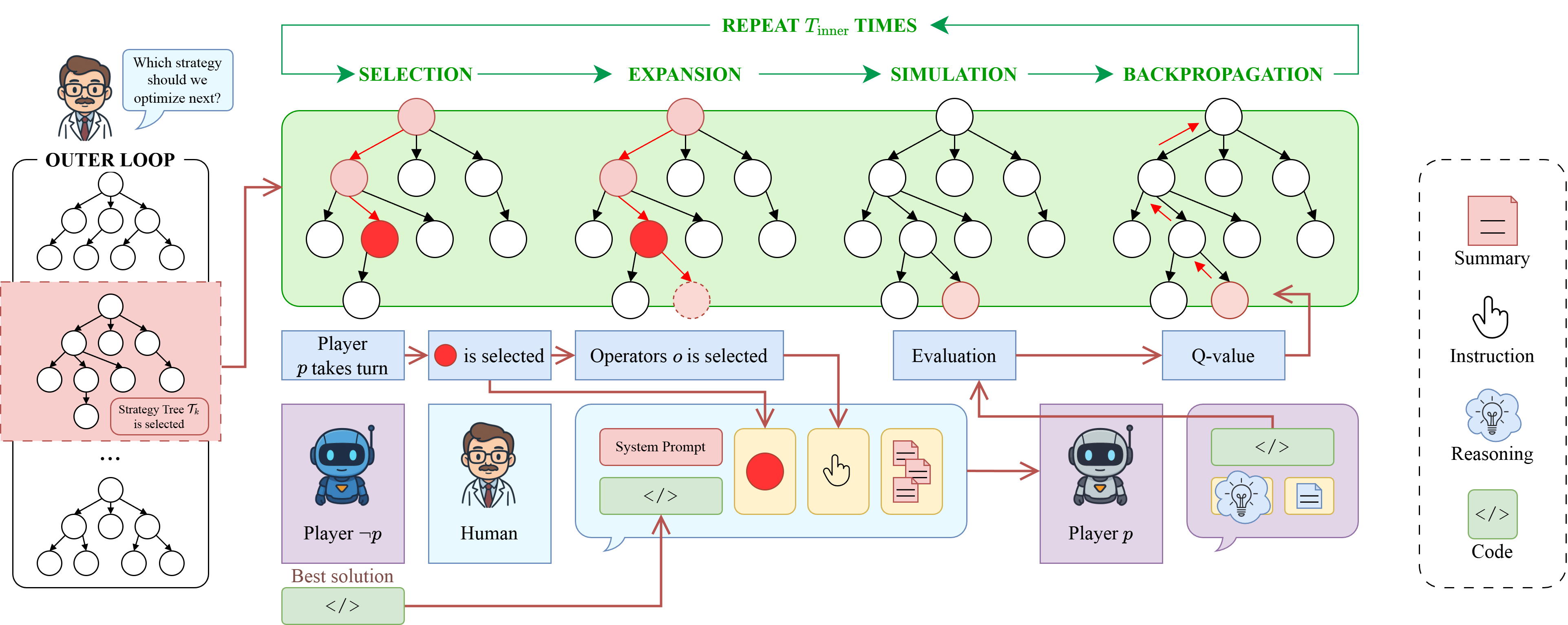}
    \caption{Overview of the component-wise competition framework. \textbf{Left}: The outer controller selects a strategy tree \(\mathcal{T}_k\) to optimize in each iteration. \textbf{Right}: The selected tree is improved via a two-player CMCTS, where agents alternate turns using one operator. Each move prompts the LLM with contextualized information about the current and opponent implementations, as well as prior history. Generated code is evaluated and backpropagated through the tree based on a Q-value that accounts for both absolute and relative improvements. The best solution is retained for potential system-level baseline updates.}
    \label{fig:overall}
\end{figure*}

\subsection{Architectural Overview}

MOTIF adopts a two-round optimization process that reflects a natural progression from local specialization to global integration. The goal is to gradually increase the cognitive burden placed on the agents, allowing them to first perform focused improvements under simplified context before advancing to more complex, system-aware reasoning.

\paragraph{Component-wise Competition.} 
In the first phase, the solver is decomposed into a set of individual strategies \(\{\pi_1, \pi_2, \dots, \pi_K\}\), each optimized independently. A separate competitive tree is constructed for each strategy, where two agents alternately propose revisions to a single component at a time. During this phase, agents operate with partial context: they have access only to the implementation and performance of the target strategy, its dynamic baseline, and the opponent’s proposal.

\paragraph{System-aware Refinement.}
Once all components have been optimized in isolation, the second phase begins. Here, agents revisit each strategy sequentially, but now with access to the full system configuration. At each turn, a player proposes a modification to one strategy while observing the entire combination of current implementations. A fixed global baseline is used during the optimization of each component, ensuring fairness and stability across turns. This phase encourages the emergence of synergistic adaptations, where the effectiveness of one strategy depends on how well it integrates with others.

\paragraph{Turn-based Dual-agent Game.}
Throughout both phases, the optimization proceeds as a turn-based game. At each turn, a player selects an operator to generate a new implementation, aiming not only to beat the baseline but also to outperform the opponent. This structure fosters both adversarial behavior—by incentivizing relative gains—and cooperative dynamics—by reinforcing global cost reduction. The continuous interplay between agents creates a rich and adaptive optimization trajectory that evolves from competitive local improvements to system-level coherence.

\subsection{Component-wise Competition}

\paragraph{Outer Controller and Strategy Selection.}
In the first phase, MOTIF maintains a separate tree \(\mathcal{T}_k\) for each strategy \(\pi_k\). At each outer iteration, a controller selects one tree to optimize using a UCB-based rule:
\begin{equation}
k = \arg\max_{1 \leq k \leq K} \left( \frac{R_j}{N_j} + C_{\text{out}} \sqrt{\frac{\ln \sum_i N_i}{N_j}} \right),
\end{equation}
where \(R_j\) and \(N_j\) are the total reward and visit count of tree \(\mathcal{T}_j\), and \(C_{\text{out}}\) controls exploration.

The selected strategy is then optimized via a two-player competition. Although only one component is updated per turn, its impact is assessed globally via the system cost.

\paragraph{Node Representation.}
Each node represents a game state where $\mathcal{P}_1$ and $\mathcal{P}_2$ each hold a distinct implementation of $\pi_k$, along with their respective code, and improvement metrics.

Nodes are linked through operator-based transformations, and updated using turn-aware reward signals. Since only one player acts per turn, the structure supports asymmetric credit assignment and promotes adaptive behavior. Each \(\mathcal{T}_k\) evolves via a Competitive MCTS, as described next.

\paragraph{Competitive Monte Carlo Tree Search.}
In \gls{cmcts}, each node represents a two-player duel over a specific strategy \(\pi_k\), where both players maintain separate implementations and cost estimates. The tree expands through one of three competitive operators that guide the language model \(\mathcal{L}\) in proposing new code:

\begin{itemize}
    \item \textit{Counter} targets weaknesses in the opponent’s code, prompting the LLM to design adversarial improvements that exploit inefficiencies or limitations.
    \item \textit{Learning} encourages synthesis by integrating strengths from the opponent’s implementation into the agent’s own solution.
    \item \textit{Innovation} promotes exploration by instructing the LLM to ignore prior solutions and propose novel, potentially unconventional approaches.
\end{itemize}

Together, these operators span a rich spectrum of behaviors—from adversarial exploitation to constructive integration to exploratory invention. During search, the UCB formula is applied at the operator level to ensure systematic coverage of diverse transformation types.

As illustrated in Figure~\ref{fig:overall}, the competitive search follows a standard MCTS procedure with the following four steps:

\paragraph{1. Selection.}
The search begins at the root and follows the child nodes created by the current player. If all operators have been explored at a node, the child with the highest average Q-value is selected. Otherwise, an unexplored operator is chosen based on a UCB rule:
\begin{equation}
\text{UCB}(o,\pi) = \frac{Q(o,\pi)}{N(o,\pi)} + C_{\text{in}} \sqrt{\frac{\ln N(\pi)}{N(o,\pi)}},
\end{equation}
where \(Q(o,\pi)\) is the cumulative reward, \(N(o,\pi)\) the visit count for operator \(o\), and \(N(\pi)\) the total visits to node \(\pi\). The constant \(C_{\text{in}}\) balances exploration and exploitation.

\paragraph{2. Expansion.}
A new node is created by applying the selected operator to the current player’s implementation. Let \(p \in \{\mathcal{P}_1, \mathcal{P}_2\}\) be the active player and \(\pi^{(p)}\) their current code. The language model \(\mathcal{L}\) generates a new implementation via:
\begin{equation}
\pi' \leftarrow \mathcal{L}\left(\text{Prompt}\left(o; \left\{\pi^{(p)}, \pi^{(\neg p)}\right\};\left\{ \mathcal{H}^{(p)}, \mathcal{H}^{(\neg p)}\right\}; \mathcal{B}\right)\right),
\end{equation}
where \(o \in \{\text{counter},\, \text{learning},\, \text{innovation}\}\) is the selected operator type; \(\pi^{(p)}\), \(\pi^{(\neg p)}\) are the current implementations of the active player and their opponent, respectively; \(\mathcal{H}^{(p)}\), \(\mathcal{H}^{(\neg p)}\) are the recent move histories of each player, including summaries of past improvements and operator usage; \(\mathcal{B}\) denotes the current baseline, consisting of the reference implementation and its associated system cost.

\paragraph{3. Evaluation.}
The modified implementation \(\pi'\) is inserted into the system, replacing the corresponding strategy. The resulting cost is compared to the fixed baseline to compute the improvement \(I^{(p)}\) (\%) for the current player \(p\). The opponent’s value \(I^{(\neg p)}\) (\%) reflects their unchanged performance at the same node, inherited from the parent.

\paragraph{4. Backpropagation.}
The resulting reward is propagated upward through the tree. For player \(p\), the Q-value combines both absolute improvement and competitive gain:
\begin{equation}
Q^{(p)} = \lambda \cdot \sigma(I^{(p)}) + (1 - \lambda) \cdot \sigma(I^{(p)} - I^{(\neg p)}),
\end{equation}
where \(\sigma(x) = \frac{1}{1 + e^{-kx}}\) and \(\lambda \in [0,1]\) controls the balance. This reward updates statistics for both the visited nodes and the operator used, supporting informed selection in future iterations.

\paragraph{Dynamic Baseline.} To drive meaningful progress during optimization, we adopt a dynamic baseline mechanism. At any outer iteration, the baseline refers to the best-known implementation of a strategy—initially handcrafted or warm-started—and is updated whenever an agent produces a strictly better implementation. Each agent competes not against a static reference, but against the most recent high-performing solution from the opposite player.

\subsection{System-aware Refinement}

% After all strategies have been optimized individually, the second phase reconsiders each \(\pi_k\) in the context of the full solver. Unlike the first round, where agents operated on isolated components, the final round exposes the entire strategy combination \(\mathbf{\Pi}=(\pi_1, \dots, \pi_K)\) to both players.

% At each step, a fixed baseline combination is used, and the players alternate turns proposing improvements to a single strategy \(\pi_k\). The key distinction lies in the prompt construction: rather than focusing on local context and opponent-specific features, the prompt now includes the full system configuration, global baseline cost, and historical search traces. This enables the language model \(\mathcal{L}\) to reason about system-level dependencies, hyperparameter synergies, and global optimization behavior:
% \begin{equation}
% \pi_k' \leftarrow \mathcal{L}\left(\text{Prompt}\left(\mathbf{\Pi};\left\{ \pi_k^{(p)},\pi_k^{(\neg p)}\right\}; \left\{ \mathcal{H}^{(p)},\mathcal{H}^{(\neg p)}\right\}\right)\right).
% \end{equation}

% Each strategy is revisited in turn using this system-aware prompting, and improvements are accepted only if they yield a lower cost than both the fixed baseline and the opponent’s best result. This phase encourages cross-strategy adaptation and the discovery of synergistic configurations that are not visible when optimizing components in isolation.

In the second phase, each strategy \(\pi_k\) is re-optimized with full system visibility. Unlike the first phase, players now operate over the entire configuration \(\mathbf{\Pi} = (\pi_1, \dots, \pi_K)\), using a fixed baseline for each strategy.

The key distinction lies in the prompt construction: rather than focusing on local context and opponent-specific features, the prompt now includes the full system configuration, global baseline cost, and historical search traces. This enables the language model \(\mathcal{L}\) to reason about system-level dependencies, hyperparameter synergies, and global optimization behavior:
\begin{equation}
\pi_k' \leftarrow \mathcal{L}\left(\text{Prompt}\left(\mathbf{\Pi};\left\{ \pi_k^{(p)},\pi_k^{(\neg p)}\right\}; \left\{ \mathcal{H}^{(p)},\mathcal{H}^{(\neg p)}\right\}\right)\right).
\end{equation}

An update is accepted only if it improves upon both the baseline and the opponent’s best result, promoting coordination across components and discovering configurations not reachable through isolated optimization. See Appendix~\ref{app:second} for more details.

% \begin{figure}
%   \centering
%   \includegraphics[scale=0.6]{figures/method.pdf}
%   \caption{Hello}
%   \label{fig:my_figure}
% \end{figure}

\section{Experiments}
\label{others}

\paragraph{Settings.}  We employ the \texttt{gpt-4o-mini-2024-07-18} model for both agents throughout our experiments. This model, although not highly capable in coding tasks, was deliberately chosen for its affordability, fast inference, and support for structured outputs, which are essential for reliable parsing and downstream processing. Given its limited reasoning ability, we design a structured response format that enforces clarity in its generation process.

Specifically, we adopt a lightweight variant of chain-of-thought prompting \cite{wei2023chainofthoughtpromptingelicitsreasoning}. Each response must contain a \texttt{reasoning} field, in which the model is required to explain its thought process in no more than five concise sentences. The \texttt{code} field contains the generated Python implementation, while the \texttt{summary} field briefly describes the key changes introduced in the current turn. This tri-field schema helps us better monitor reasoning quality, code correctness, and strategic intent during each optimization step.

\paragraph{Training and Evaluation Setup.} All optimization is conducted on a lightweight training set, while final performance is measured on a held-out test set. Full details of the setup, including dataset construction and evaluation protocol, are provided in Appendix~\ref{app:data}.\footnote{Our
code is available at: github.com/HaiAu2501/MOTIF}

\subsection{Single-strategy Search}

\paragraph{Guided Local Search.} Among classical metaheuristics for combinatorial optimization, one of the most effective is \gls{gls} \cite{VOUDOURIS1999469}, which enhances local search by penalizing frequent features of poor local optima. It has demonstrated strong performance on combinatorial problems such as the \gls{tsp}. In practice, \gls{gls} is often deployed with handcrafted heuristics that guide the search toward promising regions of the solution space.

To preserve the efficiency of GLS, prior works typically optimize only a single heuristic component—usually the precomputed penalty scoring function—while keeping the surrounding search logic fixed. We adopt the same design philosophy to ensure fair comparison and fast runtime during evaluation.

As shown in Table~\ref{tab:gls}, even under this restricted single-strategy setting, our method significantly outperforms several recent LLM-based methods.

\begin{table}[t]
\centering
\caption{Average optimal gap (\%) on TSP with GLS: comparison across methods (3 runs).}
\label{tab:gls}
\renewcommand{\arraystretch}{1.2}
\resizebox{\linewidth}{!}{
\begin{tabular}{l cccc}
\toprule
\multicolumn{1}{c}{Methods} & TSP50 & TSP100 & TSP200 & TSP500 \\
\midrule
EoH & 0.0000 $\pm$ 0.0000 & 0.0012 $\pm$ 0.0009 & 0.0639 $\pm$ 0.0097 & 0.5796 $\pm$ 0.0146 \\
ReEvo & 0.0000 $\pm$ 0.0000  & 0.0335 $\pm$ 0.0449 & 0.2081 $\pm$ 0.1927 & 0.7918 $\pm$ 0.3051  \\
HSEvo & 0.0108 $\pm$ 0.0138 & 0.3095 $\pm$ 0.2363 & 1.1254 $\pm$ 0.6424 & 2.4593 $\pm$ 0.7314 \\
MCTS-AHD & 0.0000 $\pm$ 0.0000 & 0.0024 $\pm$ 0.0007 & 0.0652 $\pm$ 0.0111 & 0.5611 $\pm$ 0.0216 \\
\midrule
MOTIF & \cellcolor{gray!30}\textbf{0.0000 $\pm$ 0.0000}  & \cellcolor{gray!30}\textbf{0.0007 $\pm$ 0.0006} & \cellcolor{gray!30}\textbf{0.0610 $\pm$ 0.0100} & \cellcolor{gray!30}\textbf{0.5577 $\pm$ 0.0255} \\

\bottomrule
\end{tabular}
}
\end{table}

\subsection{Multi-strategy Search}

\paragraph{Ant Colony Optimization.}

Simulating the pheromone-based foraging behavior of ants, \gls{aco} \cite{dorigo2007ant, dorigo2018ant} has established itself as a versatile and competitive framework for solving NP-hard combinatorial problems. Classical \gls{aco} implementations consist of multiple expert-designed components, each responsible for a specific sub-behavior of the colony. In our view, these components—often treated as fixed formulas—can themselves be subject to optimization. Specifically, we target the following sub-strategies: (i) the initialization scheme for heuristic and pheromone scores, (ii) the construction policy that combines heuristic and pheromone information into a transition probability distribution, and (iii) the pheromone update rule that determines how feedback from previous solutions reinforces or weakens certain paths.

\begin{figure*}[ht]
    \centering
    \includegraphics[width=\linewidth]{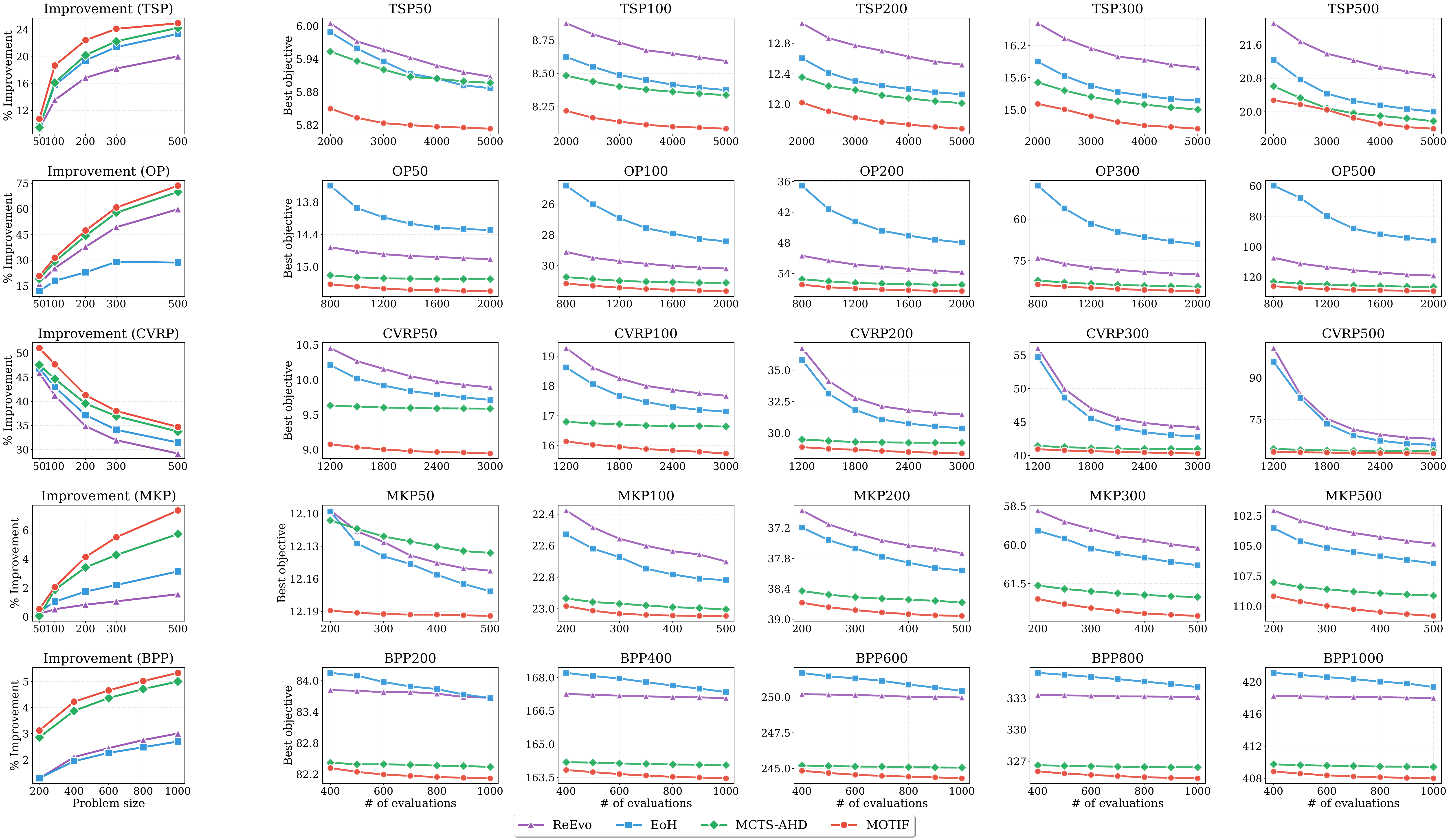}
    \caption{Comparison of AHD frameworks applied to the ACO algorithm. EoH, ReEvo, and MCTS-AHD optimize only a single strategy component (the heuristic function), while MOTIF concurrently optimizes two or three strategy components. Left: Relative performance compared to the human-designed baseline. Right: Evaluation curves showing the best objective value over time (measured by the number of evaluations), averaged over three independent runs.}
    \label{fig:aco}
\end{figure*}

Figure~\ref{fig:aco} compares the performance of our proposed MOTIF framework with several competitive baselines across five distinct optimization problems, each evaluated at five problem scales. The results demonstrate that jointly optimizing multiple strategies in ACO yields substantial performance gains over both human-designed and LLM-generated single-strategy baselines.

% \begin{table*}[t]
% \centering
% \renewcommand{\arraystretch}{1.2}
% \resizebox{\linewidth}{!}{
% \begin{tabular}{l | ccc | cccccc}
% \toprule
% Problem & Christofides &  Nearest-insertion & Nearest-greedy  & GPHH & EoH & ReEvo & HSEvo & MCTS-AHD & MOTIF\\
% \midrule
% tsp225 &  & 5.68 & - &  \\
% rat99 &  & &  & \\
% bier127 & & & & \\
% lin318 & & & & \\
% eil51 & &  & &  &\\
% d493 & &  & &  &\\
% kroB100 & &  & &  &\\
% kroC100 & &  & &  &\\
% ch130 & &  & &  &\\
% pr299 & &  & &  &\\
% fl417 & &  & &  &\\
% kroA150 & &  & &  &\\
% pr264 & &  & &  &\\
% pr226 & &  & &  &\\
% pr439 & &  & &  &\\
% \midrule
% Average & Type E & & 0.000 &  \\

% \bottomrule
% \end{tabular}
% }
% \caption{Comparison of methods on different TSP sizes.}
% \end{table*}

% \subsection{Constructive Algorithm}

\paragraph{Deconstruction-then-Repair.}

We introduce a simple yet effective constructive framework as the starting point for strategy exploration. The framework operates in three sequential stages: (i) a greedy initialization guided by an edge scoring function \(\pi_1\); (ii) a partial destruction phase that removes low-quality elements based on a badness scoring strategy \(\pi_2\); and (iii) a repair phase that incrementally reconstructs the solution using a placement selection criterion \(\pi_3\). This process reflects a natural design philosophy: producing reasonably good solutions quickly without engaging in costly search loops.

Table~\ref{tab:ga} reports the improvement margins achieved by optimizing one, two, or all three strategies in the framework. Each component contributes differently depending on the problem domain—for example, \(\pi_3\) is critical in TSP, while \(\pi_2\) plays a larger role in CVRP and BPP. Overall, jointly optimizing two or more strategies consistently outperforms the single-strategy setting, demonstrating the synergistic benefits of multi-strategy optimization.

These results also suggest that, beyond merely tuning isolated heuristics, LLMs hold promise in co-designing full algorithmic pipelines. This supports the broader vision of transforming a simple, human-sketched pipeline into a strong, domain-adapted algorithm through iterative LLM-guided search.

\begin{table*}[ht]
\centering
% $\uparrow$
\caption{Performance comparison between single-strategy and multi-strategy optimization across three combinatorial problems: TSP, CVRP, and BPP, each evaluated at various instance sizes. Results indicate percentage improvement over the human-designed baseline, averaged over 3 runs. Strategies $\pi_1,\pi_2,\pi_3$ denote initialization, deconstruction, and repair respectively. } % Notably, combinations involving multiple strategies  consistently outperform single-strategy baselines, demonstrating the advantage of coordinated multi-strategy optimization under the proposed framework.
\renewcommand{\arraystretch}{1.2}
\resizebox{\linewidth}{!}{
\begin{tabular}{c ccc ccc ccc}
\toprule
\multirow{2}{*}{Strategies} & \multicolumn{3}{c}{\textbf{TSP}} & \multicolumn{3}{c}{\textbf{CVRP}} & \multicolumn{3}{c}{\textbf{BPP}}  \\
\cmidrule(lr){2-4} \cmidrule(lr){5-7} \cmidrule(lr){8-10} 
& 50 & 100 & 200 & 50 & 100 & 200 & 100 & 200 & 300 \\
\midrule
% $\varnothing$ & 6.81 & 9.62 & 13.49 & 0.00 & 0.00 & 0.00 & 0.00 & 0.00 & 0.00\\
% \midrule
$\pi_1$ 
& 0.39 $\pm$ 0.17 & 0.81 $\pm$ 0.27 & 0.72 $\pm$ 0.45 
& 1.47 $\pm$ 0.16 & 2.96 $\pm$ 0.25 & 2.30 $\pm$ 0.39
& 3.17 $\pm$ 0.00 & 4.82 $\pm$ 0.00 & 4.77 $\pm$ 0.00
\\ 

$\pi_2$ 
& 3.13 $\pm$ 0.05 & 6.24 $\pm$ 0.21 & 8.18 $\pm$ 0.26 
& 4.27 $\pm$ 0.24 & 7.04 $\pm$ 0.73 & 6.85 $\pm$ 0.75
& 23.19 $\pm$ 0.05 & 24.42 $\pm$ 0.02 & 25.00 $\pm$ 0.00
\\

$\pi_3$ 
& \cellcolor{gray!30}\textbf{3.88 $\pm$ 0.04} & 7.91 $\pm$ 0.01 & 11.35 $\pm$ 0.03 
& 6.34 $\pm$ 0.08 & 5.07 $\pm$ 0.02 & 4.28 $\pm$ 0.03
& 12.70 $\pm$ 7.51 & 12.84 $\pm$ 7.15 & 12.15 $\pm$ 7.49
\\ 

\midrule
$(\pi_1,\pi_2)$ 
& 2.58 $\pm$ 0.50 & 5.04 $\pm$ 0.78 & 5.84 $\pm$ 1.26 
& 6.06 $\pm$ 1.64 & 7.94 $\pm$ 2.71 & 8.85 $\pm$ 2.64
& 23.15 $\pm$ 0.15 & 24.40 $\pm$ 0.06 & 24.98 $\pm$ 0.02
\\

$(\pi_2,\pi_3)$ & 3.83 $\pm$ 0.16 & \cellcolor{gray!30}\textbf{7.97 $\pm$ 0.14} & 11.59 $\pm$ 0.07
& 10.64 $\pm$ 0.74 & 12.31 $\pm$ 0.57 & 11.71 $\pm$ 1.03 
& 23.05 $\pm$ 0.11 & 24.32 $\pm$ 0.15 & 24.89 $\pm$ 0.15
\\
\midrule

$(\pi_1,\pi_2,\pi_3)$ 
& \cellcolor{gray!30}\textbf{3.88 $\pm$ 0.04} & 7.96 $\pm$ 0.03 & \cellcolor{gray!30}\textbf{11.65 $\pm$ 0.05}
& \cellcolor{gray!30}\textbf{10.98 $\pm$ 1.64} & \cellcolor{gray!30}\textbf{12.84 $\pm$ 3.17} & \cellcolor{gray!30}\textbf{13.06 $\pm$ 3.67}
& \cellcolor{gray!30}\textbf{23.94 $\pm$ 0.55} & \cellcolor{gray!30}\textbf{25.02 $\pm$ 0.41} & \cellcolor{gray!30}\textbf{25.41 $\pm$ 0.26}
\\
\bottomrule
\end{tabular}
}

\label{tab:ga}
\end{table*}

\subsection{Convergence and Diversity Analysis}
% \vspace{-5 pt}
\begin{figure}[t]
    \centering
    \includegraphics[width=\columnwidth]{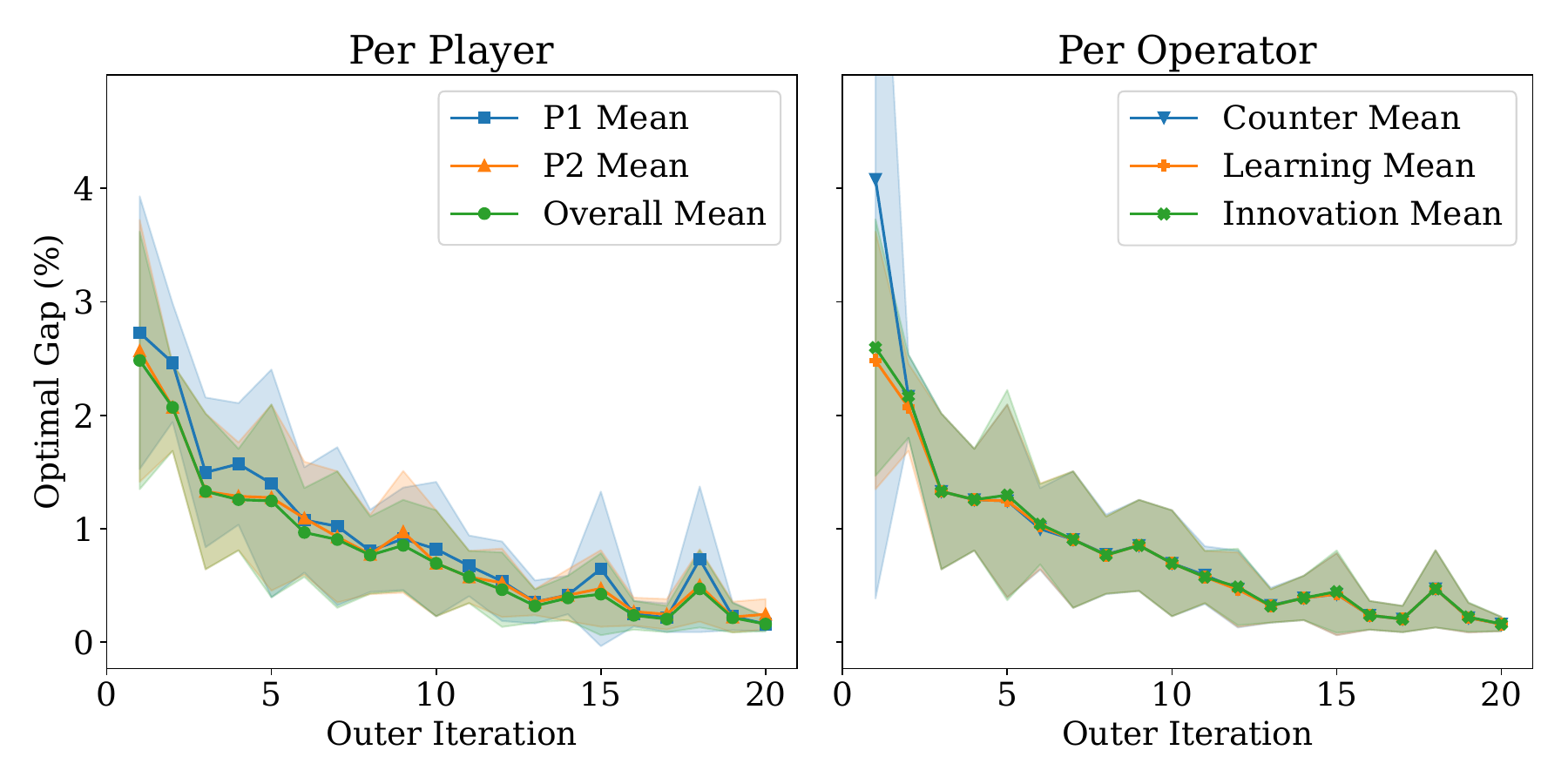}
    \caption{Convergence behavior of the MOTIF framework during training, averaged over five independent runs. \textbf{Left:} Best optimality gap achieved at each outer iteration, shown separately for Player 1 (P1), Player 2 (P2), and the overall best. \textbf{Right:} Performance breakdown by operator type.}
    \label{fig:conv}
\end{figure}

Figure~\ref{fig:conv} plots the best optimality gap achieved at each outer iteration across five independent training runs. While occasional regressions are observed—due to the exploratory nature of operator moves—the overall trend is clearly downward, indicating steady convergence. 

Interestingly, both players exhibit closely aligned performance curves throughout training, suggesting a dynamic equilibrium in which competitive pressure drives mutual improvement. Furthermore, as shown on the right, all three operators—\textit{counter}, \textit{learning}, and \textit{innovation}—demonstrate similar convergence profiles, reflecting the robustness and adaptability of our operator design across varying strategic contexts.

% \noindent
% \begin{minipage}{0.48\columnwidth}
% \includegraphics[width=\columnwidth]{figures/Fig5.pdf}
% \end{minipage}
% \hfill
% \begin{minipage}{0.48\columnwidth}
% \centering
% \renewcommand{\arraystretch}{1.2}
% \resizebox{\columnwidth}{!}{%
% \begin{tabular}{ccc}
% \toprule
% Operator & Novelty & Intra-Diversity \\
% \midrule
% Counter & 1 & \\
% Learning & 2 & \\
% Innovation & 3 & \\
% \bottomrule
% \end{tabular}}
% \end{minipage}

\begin{table}[h]
    \caption{Diversity and success analysis for each operator. Success rate measures the proportion of generated implementations that improve upon the current baseline. Novelty score captures the average semantic distance to other operators’ outputs within the same strategy. Silhouette score quantifies intra-operator cohesion and inter-operator separation in the embedding space.
All metrics are averaged over five independent runs.}
    \label{tab:div}
\centering
\renewcommand{\arraystretch}{1.2}
\resizebox{\columnwidth}{!}{%
\begin{tabular}{lccc}
\toprule
\multicolumn{1}{c}{Operator} & Success rate $(\uparrow)$ & Novelty score $(\uparrow)$ & Silhouette score $(\uparrow)$ \\
\midrule
Counter & 93 $\pm$ 2 \% & 0.0136 $\pm$ 0.0067 & 0.5121 $\pm$ 0.0208 \\
Learning & 92 $\pm$ 2 \% & 0.0118 $\pm$ 0.0054 & \cellcolor{gray!30}\textbf{0.5325 $\pm$ 0.0300} \\
Innovation & 97 $\pm$ 1 \% & \cellcolor{gray!30}\textbf{0.0175 $\pm$ 0.0110} & 0.4793 $\pm$ 0.0240 \\
\bottomrule
\end{tabular}}
\end{table}

Another question in operator design is whether LLM-based strategies can break free from conventional coding patterns and exhibit genuinely novel behavior. To examine this, we conduct a semantic diversity analysis of the generated implementations. Specifically, we compute two metrics—\emph{novelty} and \emph{silhouette score}—based on code embeddings to assess how distinct and well-separated each operator’s outputs are. Formal definitions and computation details of these metrics are provided in Appendix~\ref{app:metrics}.

Table~\ref{tab:div} compares the three operators across success rate, novelty, and silhouette score. \textit{Innovation} exhibits the highest novelty, reflecting its broad exploration of new code regions. However, it has the lowest silhouette score, suggesting its outputs are scattered and lack internal consistency. \textit{Counter} achieves moderate novelty and silhouette, indicating a balanced behavior—exploring new directions while maintaining some cohesion. \textit{Learning} ranks lowest in novelty but highest in silhouette, showing that it tends to exploit familiar patterns with consistent and stable outputs.

These trends align with the intended design: \textit{innovation} promotes diversity, \textit{learning} refines known ideas, and \textit{counter} responds adaptively to the opponent.

\subsection{Ablation Study}

Table~\ref{tab:abalations} reports the impact of removing various components on the final optimal gap (lower is better) for two solvers: ACO and GLS. Among three components, removing the dynamic baseline causes the most severe performance drop, especially in the test phase—indicating that without continual baseline updates, the system lacks incentive to improve and tends to stagnate.

Disabling reasoning also leads to a notable degradation, highlighting its importance in enabling the model to reflect on prior failures and generate meaningful revisions.

\begin{table}[h]
\centering
\caption{Ablations on components, prompting, and operators. Numbers denote optimality gap (\%, lower is better), averaged over 3 runs. Results are reported under a small evaluation budget, using TSP50 for training and TSP100 for testing, to highlight performance differences more clearly.}
\renewcommand{\arraystretch}{1.2}
\resizebox{\columnwidth}{!}{
\begin{tabular}{l cc cc c}
\toprule
\multicolumn{1}{c}{\multirow{2}{*}{Methods}} & \multicolumn{2}{c}{ACO} & \multicolumn{2}{c}{GLS}   \\
\cmidrule(lr){2-3} \cmidrule(lr){4-5}
& Train & Test & Train & Test \\
\midrule
w/o Outer Controller
& 0.74 $\pm$ 0.39 & 5.61 $\pm$ 0.86
& 0.81 $\pm$ 0.14 & 2.88 $\pm$ 0.11
\\
w/o Dynamic Baseline
& 1.55 $\pm$ 0.26 & 9.50 $\pm$ 4.24 
& 0.62 $\pm$ 0.20 & 2.94 $\pm$ 0.18 
\\
w/o Final Round 
& 1.26 $\pm$ 0.52 & 5.72 $\pm$ 0.78 
& 0.39 $\pm$ 0.13 & 2.82 $\pm$ 0.06
\\
\midrule
w/o Reasoning 
& 1.63 $\pm$ 0.94 & 7.88 $\pm$ 3.76 
& 0.80 $\pm$ 0.23 & 3.05 $\pm$ 0.04 
\\ 
w/o Active Player's History 
& 0.82 $\pm$ 0.63 & 6.06 $\pm$ 1.27 
& 0.53 $\pm$ 0.07 & 2.97 $\pm$ 0.04
\\ 
w/o Opponent's History 
& 0.97 $\pm$ 0.50 & 5.58 $\pm$ 1.31 
& 0.48 $\pm$ 0.00 & 2.98 $\pm$ 0.13
\\
\midrule
w/o Counter 
& 1.42 $\pm$ 0.07 & 6.71 $\pm$ 0.52 
& 0.64 $\pm$ 0.19 & 3.20 $\pm$ 0.12
\\
w/o Learning 
& 0.91 $\pm$ 0.32 & 9.59 $\pm$ 5.74 
& 0.62 $\pm$ 0.02 & 3.00 $\pm$ 0.12
\\
w/o Innovation 
& 1.73 $\pm$ 0.31 & 6.65 $\pm$ 0.80
& 0.42 $\pm$ 0.09 & 2.86 $\pm$ 0.00
\\
\midrule
MOTIF (original) 
& 0.80 $\pm$ 0.28 & \cellcolor{gray!30}\textbf{5.21 $\pm$ 0.35}
& 0.34 $\pm$ 0.19 & \cellcolor{gray!30}\textbf{2.73 $\pm$ 0.05}
\\ 
\bottomrule
\end{tabular}
}
\label{tab:abalations}
\end{table}
\section{Conclusion}
We introduced a two-phase competitive optimization framework that leverages turn-based interactions between LLM agents to improve multi-strategy solvers. Through dynamic baselines, self-play prompting, and system-aware refinement, our method consistently outperforms prior approaches across diverse combinatorial problems. The results highlight the importance of both adversarial pressure and structured cooperation in driving algorithmic innovation.

\section*{Acknowledgement}
This research was funded by Hanoi University of Science and Technology under project code T2024-PC-038.

% \begin{ack}
% Use unnumbered first level headings for the acknowledgments. All acknowledgments

% Do {\bf not} include this section in the anonymized submission, only in the final paper. You can use the \texttt{ack} environment provided in the style file to automatically hide this section in the anonymized submission.
% \end{ack}

\bibliography{bibliography}

% \section*{References}

% References follow the acknowledgments in the camera-ready paper. Use unnumbered first-level heading for
% the references. Any choice of citation style is acceptable as long as you are
% consistent. It is permissible to reduce the font size to \verb+small+ (9 point)
% when listing the references.
% Note that the Reference section does not count towards the page limit.
% \medskip

% {
% \small

% [1] Alexander, J.A.\ \& Mozer, M.C.\ (1995) Template-based algorithms for
% connectionist rule extraction. In G.\ Tesauro, D.S.\ Touretzky and T.K.\ Leen
% (eds.), {\it Advances in Neural Information Processing Systems 7},
% pp.\ 609--616. Cambridge, MA: MIT Press.

% [2] Bower, J.M.\ \& Beeman, D.\ (1995) {\it The Book of GENESIS: Exploring
%   Realistic Neural Models with the GEneral NEural SImulation System.}  New York:
% TELOS/Springer--Verlag.

% [3] Hasselmo, M.E., Schnell, E.\ \& Barkai, E.\ (1995) Dynamics of learning and
% recall at excitatory recurrent synapses and cholinergic modulation in rat
% hippocampal region CA3. {\it Journal of Neuroscience} {\bf 15}(7):5249-5262.
% }

%%%%%%%%%%%%%%%%%%%%%%%%%%%%%%%%%%%%%%%%%%%%%%%%%%%%%%%%%%%%

\cleardoublepage
\newpage
\appendix
\onecolumn
% \renewcommand{\thesection}{\Alph{section}}

% \section*{Appendix: Table of Contents}
% \vspace{1em}

% \begin{itemize}[leftmargin=*, label={}]
%     \item \textbf{A. Related Works} \dotfill~\pageref{appendix: related}
%     \begin{itemize}[leftmargin=*, label={}]
%         \item A.1 \quad Hyper-heuristic Frameworks \dotfill~\pageref{appendix: hh}
%         \item A.2 \quad Evolutionary and LLM-Driven Heuristic Design \dotfill~\pageref{appendix: EA-LLM}
%         \item A.3 \quad Monte Carlo Tree Search for Heuristic and Strategy Search \dotfill~\pageref{appendix: MCTS-LLM}
%         \item A.4 \quad Self-play and Adversarial Learning in LLMs \dotfill~\pageref{appendix: Self-play}
%     \end{itemize}

%     \item \textbf{B \quad Problem Definitions} 
%     \dotfill~\pageref{appendix: problem_def}
%     \begin{itemize}[leftmargin=*, label={}]
%         \item B.1 \quad Benchmark Problems \dotfill~\pageref{appendix: benchmark problems}
%         \item B.2 \quad Benchmark Algorithms \dotfill~\pageref{appendix: benchmark algo}
%     \end{itemize}

%     \item \textbf{C \quad Methodology Details} \dotfill~\pageref{appendix: Method}
%     \item \textbf{D \quad Experiment Details} \dotfill~\pageref{appendix: experiment}
% \end{itemize}

\section{Related Works}
\label{appendix: related}

\subsection{Early Hyper-heuristic Frameworks}

\paragraph{General Hyper-heuristics.}
Hyper-heuristics, often referred to as \acrfull{ahd}, are high-level search frameworks that operate over a space of low-level heuristics instead of directly modifying problem solutions \cite{zhang2023survey}. The primary motivation behind this paradigm is to enhance generality and transferability across combinatorial optimization problems. By abstracting away domain-specific details, hyper-heuristics decouple the problem-solving process from the manual crafting of individual heuristics. Foundational work by \citet{burke} categorized these frameworks into two principal classes: selection hyper-heuristics, which dynamically choose from a set of predefined heuristics, and generation hyper-heuristics, which synthesize new heuristics during the search. Their subsequent survey \citep{burke2013} documented the rising maturity of the field, especially in classical domains such as scheduling, bin packing, and routing.

\paragraph{Hyper-heuristics via Genetic Programming.}
A notable instantiation of generation-based hyper-heuristics is Genetic Programming (GP) \cite{koza1994genetic}, where heuristics are evolved as symbolic expressions. The two-layer architecture, comprising a domain-agnostic controller and domain-specific heuristic components, naturally lends itself to GP-based designs. \citet{burke2009exploring} explored how GP could be used to evolve compositional heuristic logic, while \citet{burke2010genetic} demonstrated the feasibility of this approach in practical applications such as two-dimensional strip packing. These systems construct heuristics as executable programs, enabling fine-grained adaptation and a rich space of search behaviors. This line of work laid the groundwork for modern approaches that further extend the heuristic search space with neural or language-based models—but those advancements are discussed in later sections.

\subsection{Evolutionary and LLM-Driven Heuristic Design}

\paragraph{Evolutionary Computation.}
Before the advent of LLMs, evolutionary computation had already established itself as a powerful paradigm for AHD \cite{zhang2020evolving, zhang2021surrogate}. Inspired by principles of natural evolution, these methods evolve a population of candidate heuristics through iterative application of operators such as mutation (local changes to code), crossover (combination of two heuristics), and selection (survival of the fittest). Genetic programming in particular allowed symbolic expressions representing heuristic logic to be evolved over time, forming the conceptual basis for later work that combines these ideas with modern neural models. Crucially, this population-based, variation-and-selection mechanism provides a flexible search backbone, one that LLMs can now plug into as heuristic generators or mutators.

\paragraph{LLM-Driven Heuristic Design.}
Recent work has explored how large language models can be integrated into evolutionary frameworks to automate the design of heuristics. These systems typically prompt LLMs to generate candidate routines—often as executable Python functions—which are then evaluated and selected based on performance. In EoH~\cite{liu2024evolution}, GPT-4 is used to synthesize scoring functions for combinatorial problems, guided by tournament selection to evolve increasingly effective heuristics over generations. ReEvo~\cite{ye2024reevo} extends this setup by encouraging the LLM to reflect on its own outputs through prompt chaining and post-hoc critique, improving generation quality via self-evaluation.

Beyond individual refinement, HSEvo~\cite{dat2025hsevo} incorporates LLMs into a harmony search framework, maintaining a diverse population of heuristics akin to musical motifs. It emphasizes not only optimization but also population diversity, leveraging the generative breadth of LLMs to explore complementary algorithmic structures. 

This fusion of LLMs with evolutionary computation marks a shift in how algorithm design is conceptualized: from crafting static heuristics to evolving dynamic, language-based solvers that can interact with selection, critique, and competition.

\subsection{Monte Carlo Tree Search in Complex Decision-making Problems}

\paragraph{Monte Carlo Tree Search.}
Monte Carlo Tree Search (MCTS) has emerged as a powerful paradigm for navigating large, structured decision spaces where exhaustive search is intractable. Rooted in sequential decision theory, classical MCTS—first formalized by \citet{mcts} and comprehensively surveyed by \citet{browne2012survey}—follows a four-stage loop: \textit{selection}, \textit{expansion}, \textit{simulation}, and \textit{backpropagation}. Each node in the tree represents a partial decision trajectory, and the use of Upper Confidence Bounds (UCB) enables principled balancing between exploration of uncertain paths and exploitation of known promising ones.

MCTS gained widespread recognition through its role in AlphaGo~\cite{silver2016mastering}, where it served as the backbone for combining deep neural networks with search-based planning. In this context, MCTS operated not just as a passive evaluator, but as an active controller capable of orchestrating deep policy and value networks for superhuman performance in the game of Go. Since then, MCTS has found applications in planning, robotics \cite{dam2022monte}, theorem proving \cite{xindeepseek}, program synthesis, and most recently, language modeling.

\paragraph{LLM-Augmented Tree Search.}
A recent direction integrates MCTS with large language models (LLMs) to support structured reasoning and decision-making. The Tree-of-Thoughts framework~\cite{yao2023tree} treats LLMs as thought generators, where each node in the tree corresponds to an intermediate reasoning step (a ``thought'') toward solving a problem. MCTS is used to explore these thoughts hierarchically, evaluating partial reasoning paths and selecting which branches to expand further. This mechanism moves beyond greedy prompting or single-shot decoding by enabling deliberate, self-corrective exploration over multi-step reasoning trajectories.

This idea has since been adapted for algorithm discovery. For example, MCTS-AHD~\cite{zheng2025monte} applies progressive widening—a technique that defers branching until sufficient evidence is accumulated—to explore a search tree of heuristic strategies generated via LLM prompts. Each node in the tree represents a partial or complete heuristic configuration, and new child nodes are instantiated by prompting an LLM to revise or extend the current implementation. Selection is guided by performance metrics, and backpropagation enables learning from both successes and failures. This combination transforms MCTS into a dynamic controller over language model outputs, selectively steering generation toward effective algorithmic behaviors.

Compared to evolutionary approaches that operate on flat populations of heuristics, MCTS imposes a hierarchical structure on the search space, allowing better reuse of promising sub-strategies and more targeted exploration. Progressive widening further mitigates the risk of combinatorial explosion by limiting branching in early stages. Together, these properties make MCTS particularly well-suited for LLM-based heuristic design, where the space of possible routines is both vast and costly to evaluate. By fusing the generative flexibility of LLMs with the selective discipline of MCTS, these methods achieve a scalable and adaptive framework for navigating algorithmic design spaces.

\subsection{Self-play and Adversarial Learning in LLMs}

\paragraph{Self-play.}
Recent work has shown that self-play—where language models act as interacting agents—can lead to significant gains in reasoning quality, robustness, and decision-making. Inspired by game-theoretic dynamics, these methods structure LLM interaction through alternating roles such as attacker–defender, proposer–critic, or buyer–seller, enabling emergent behavior that improves model output beyond static prompting. For instance, SPAG~\cite{cheng2025selfplayingadversariallanguagegame} frames reasoning as a two-player adversarial game in which an attacker introduces misleading arguments and a defender must debunk them, improving the model’s capacity for deception detection. Similarly, the CDG framework~\cite{wangimproving} trains a pair of prover and critic agents to iteratively expose and correct reasoning flaws, leading to more logically sound outputs.

Other work explores negotiation and cooperation in self-play. Fu et al.~\cite{fu2023improving} design an in-context feedback loop in which LLMs simulate negotiation roles (e.g., buyer vs. seller) and revise their proposals based on AI-generated critiques, resulting in improved utility and coherence in generated dialogues. These methods typically rely on prompt structuring, role conditioning, and iterative critique to create a feedback-rich interaction landscape—one that mirrors social or argumentative reasoning among humans.

\paragraph{Adversarial Learning.} 
Adversarial learning in LLMs also intersects with verbal reinforcement learning and reflective prompting. For example, Reflexion~\cite{shinn2023reflexion} guides the model to reflect on its own past errors and learn through verbalized feedback, strengthening multi-step reasoning capabilities. Although these studies focus primarily on tasks like logic puzzles, negotiation, and QA, their results suggest that interactive multi-agent dynamics offer a powerful foundation for tasks that benefit from critique, revision, or competition. Taken together, these approaches illustrate how adversarial pressure and structured self-interaction can improve LLM outputs by introducing feedback loops, multiple perspectives, and dynamic behavior adjustment—offering a foundation for more sophisticated systems that go beyond static prompting or single-agent search.

\section{Problem Definitions}
\label{appendix: problem_def}
\subsection{Benchmark Problems}
\label{appendix: benchmark problems}

\paragraph{Traveling Salesman Problem.} 
The Traveling Salesman Problem (TSP) is defined on a complete weighted graph \( G = (V, E) \), where \( V = \{v_1, v_2, \dots, v_n\} \) is a set of cities and \( c(i, j) \in \mathbb{R}_{\geq 0} \) denotes the cost of traveling from city \( v_i \) to city \( v_j \). The goal is to find a permutation \( \sigma = (\sigma_1, \sigma_2, \dots, \sigma_n) \) of the cities that minimizes the total travel cost of a closed tour visiting each city exactly once. The objective is given by:
\begin{equation}
\min_{\sigma \in \mathcal{S}_n} \left( \sum_{i=1}^{n-1} c(\sigma_i, \sigma_{i+1}) + c(\sigma_n, \sigma_1) \right),
\end{equation}
where \( \mathcal{S}_n \) denotes the set of all permutations of \( n \) elements. The optimal solution defines the shortest Hamiltonian cycle over the graph.

\paragraph{Capacitated Vehicle Routing Problem.} 
The Capacitated Vehicle Routing Problem (CVRP) extends the TSP by introducing multiple vehicles that serve customers under capacity constraints. Let \( G = (V, E) \) be a complete undirected graph where \( V = \{v_0, v_1, \dots, v_n\} \) consists of a depot node \( v_0 \) and \( n \) customer nodes, and let \( c(i, j) \in \mathbb{R}_{\geq 0} \) denote the travel cost between nodes \( v_i \) and \( v_j \). Each customer \( v_i \) has a demand \( d_i > 0 \), and all vehicles are identical with a fixed capacity \( Q \). A route is a sequence \( r = (v_0, v_{i_1}, \dots, v_{i_m}, v_0) \) in which a single vehicle departs from the depot, serves a subset of customers, and returns to the depot. The cost of such a route is given by:
\begin{equation}
\mathrm{cost}(r) = \sum_{t=0}^{m} c(v_{i_t}, v_{i_{t+1}}),
\end{equation}
where \( v_{i_0} = v_0 \) and \( v_{i_{m+1}} = v_0 \). Each route must satisfy the capacity constraint
\begin{equation}
\sum_{j=1}^m d_{i_j} \leq Q,
\end{equation}
ensuring that the total demand served does not exceed the vehicle's capacity. The full solution consists of a set of such routes \( R = \{r_1, r_2, \dots, r_k\} \) covering all customers exactly once, and the overall objective is to minimize the total routing cost:
\begin{equation}
\sum_{r \in R} \mathrm{cost}(r).
\end{equation}

\paragraph{Multiple Knapsack Problem.} 
The Multiple Knapsack Problem (MKP) is a classic combinatorial optimization problem where a set of \( n \) items must be assigned to \( m \) distinct knapsacks with limited capacities. Each item \( i \in \{1, \dots, n\} \) has a profit \( p_i \in \mathbb{R}_{>0} \) and weight \( w_i \in \mathbb{R}_{>0} \), and each knapsack \( j \in \{1, \dots, m\} \) has a capacity \( C_j \in \mathbb{R}_{>0} \). The goal is to assign each item to at most one knapsack such that the total profit is maximized and no knapsack exceeds its capacity. Let \( x_{ij} \in \{0,1\} \) be a binary decision variable indicating whether item \( i \) is placed in knapsack \( j \). The optimization problem can be written as:
\begin{equation}
\max \sum_{j=1}^m \sum_{i=1}^n p_i x_{ij},
\end{equation}
subject to the capacity constraints
\begin{equation}
\sum_{i=1}^n w_i x_{ij} \leq C_j \quad \forall j \in \{1, \dots, m\},
\end{equation}
and assignment constraints
\begin{equation}
\sum_{j=1}^m x_{ij} \leq 1 \quad \forall i \in \{1, \dots, n\}.
\end{equation}
The MKP captures important allocation scenarios where resources are limited and item assignments are exclusive.

\paragraph{Orienteering Problem.}
The Orienteering Problem (OP) models route planning under a limited travel budget, balancing reward collection and cost. Given a graph \( G = (V, E) \) with non-negative edge costs \( c(i, j) \) for all \( (i, j) \in E \), a reward \( r_i \geq 0 \) at each node \( v_i \in V \), a start node \( v_s \in V \), an end node \( v_t \in V \), and a maximum travel budget \( B > 0 \), the objective is to find a path \( P = (v_s, \dots, v_t) \) that visits a subset of nodes such that the total reward is maximized and the total travel cost does not exceed \( B \). Let \( \mathcal{P}_{s,t} \) be the set of all feasible paths from \( v_s \) to \( v_t \) within budget. The objective is to solve:
\begin{equation}
\max_{P \in \mathcal{P}_{s,t}} \sum_{v_i \in P} r_i \quad \text{subject to} \quad \sum_{(v_i, v_j) \in P} c(i, j) \leq B.
\end{equation}
Unlike TSP or CVRP, not all nodes must be visited; instead, the challenge lies in selecting the most rewarding subset of nodes reachable within the given travel constraint.

\paragraph{Bin Packing Problem.}
The Bin Packing Problem (BPP) involves packing a set of items into the minimum number of identical bins, each with fixed capacity. Formally, let there be \( n \) items, where each item \( i \in \{1, \dots, n\} \) has a size \( s_i \in (0, 1] \), and let each bin have capacity 1. The goal is to assign each item to a bin such that the total size of items in any bin does not exceed its capacity, while minimizing the number of bins used. Let \( x_{ij} \in \{0,1\} \) indicate whether item \( i \) is placed in bin \( j \), and let \( y_j \in \{0,1\} \) indicate whether bin \( j \) is used. Assuming an upper bound \( m \geq n \) on the number of bins, the problem can be formulated as:
\begin{equation}
\min \sum_{j=1}^{m} y_j
\end{equation}
subject to:
\begin{equation}
\sum_{i=1}^{n} s_i x_{ij} \leq y_j \quad \forall j \in \{1, \dots, m\},
\end{equation}
\begin{equation}
\sum_{j=1}^{m} x_{ij} = 1 \quad \forall i \in \{1, \dots, n\},
\end{equation}
\begin{equation}
x_{ij} \in \{0, 1\}, \quad y_j \in \{0, 1\}.
\end{equation}
This formulation ensures that each item is assigned to exactly one bin, and bins are only counted if they are actually used. BPP is a fundamental NP-hard problem with applications in logistics, memory allocation, and resource scheduling.

\subsection{Benchmark Algorithms}
\label{appendix: benchmark algo}
\paragraph{Guided Local Search.} Guided Local Search (GLS) is a well-established metaheuristic designed to escape local optima by penalizing frequently occurring features in poor-quality solutions. Introduced by \citet{VOUDOURIS1999469}, GLS enhances standard local search by augmenting the objective function with feature-based penalties, thereby encouraging diversification while maintaining efficiency. It has demonstrated strong performance on various combinatorial problems, particularly the TSP, and has been used in hybrid solvers such as GLS-PR for Vehicle Routing Problems (VRP) \cite{shaw1997new, shaw1998using} and more recently in neural-guided contexts \cite{hudson2022graphneuralnetworkguided, sui2024neuralgls}.

In our benchmark, we follow the common setup used in these prior works: the main local search logic is fixed, and only the scoring function (used for feature penalization) is subject to optimization. This ensures fair comparison across methods while isolating the effect of LLM-based design.

% \begin{algorithm}[H]
% \caption{GLS Procedure for TSP}
% \begin{algorithmic}[1]
% \STATE \textbf{Input:} Distance matrix $D$, \#perturbation moves $M$, \#iterations $T$
% \STATE \colorbox{gray!20}{\textit{\textbf{Main strategy:}} $G \leftarrow$ \textsc{GenerateGuideMatrix}($D$)} 
% \STATE $P \leftarrow \mathbf{0}$ $\quad$ \COMMENT{Penalty matrix}
% \STATE $\sigma_\text{best} \leftarrow$ \textsc{NearestNeighbor}$(D)$
% \STATE \textsc{LocalSearch}$(D, \sigma_\text{best})$ $\quad$ \COMMENT{Modifies $\sigma_\text{best}$ in-place using 2-opt and relocate}
% \STATE $c_\text{best} \leftarrow \textsc{Cost}(D, \sigma_\text{best})$
% \STATE $k \leftarrow 0.1 \cdot c_\text{best} / n$ $\quad$ \COMMENT{$n$: number of cities}
% \STATE $\sigma_\text{cur} \leftarrow \sigma_\text{best}$
% \FOR{$t = 1$ to $T$}
%   \FOR{$m = 1$ to $M$}
%     \STATE $(i^*, j^*) \leftarrow \arg\max_{(i,j) \in \sigma_\text{cur}} \frac{G_{ij}}{1 + P_{ij}}$
%     \STATE $P_{i^*j^*} \leftarrow P_{i^*j^*} + 1$
%     \STATE $\hat{D} \leftarrow D + k \cdot P$
%     \STATE \textsc{LocalSearch}$(\hat{D}, \sigma_\text{cur})$ around $(i^*, j^*)$
%   \ENDFOR
%   \STATE \textsc{LocalSearch}$(D, \sigma_\text{cur})$
%   \IF{$\textsc{Cost}(D, \sigma_\text{cur}) < c_\text{best}$}
%     \STATE $\sigma_\text{best} \leftarrow \sigma_\text{cur}$
%     \STATE $c_\text{best} \leftarrow \textsc{Cost}(D, \sigma_\text{cur})$
%   \ENDIF
% \ENDFOR
% \STATE \textbf{Return:} $\sigma_\text{best}$
% \end{algorithmic}
% \end{algorithm}
\begin{algorithm}[H]
\caption{GLS Procedure for TSP}
\begin{algorithmic}[1]
\STATE \textbf{Input:} distance matrix $D$, \#perturbation moves $M$, \#iterations $T$
\STATE \colorbox{gray!20}{\textit{\textbf{Main strategy:}} $G \gets \textsc{BuildGuideMatrix}(D)$}
\STATE $P \gets \mathbf{0}, \sigma_{\text{best}} \gets \textsc{ImproveLocal}(\textsc{NearestNeighbor}(D), D)$
\FOR{$t=1..T$}
  \FOR{$m=1..M$}
    \STATE $e \gets \textsc{SelectEdgeToPenalize}(\sigma, G, P)$
    \STATE $P \gets \textsc{Penalize}(P, e)$
    \STATE $\sigma \gets \textsc{ImproveLocal}(\sigma, \textsc{AugmentCosts}(D,P))$
  \ENDFOR
  \STATE $\sigma \gets \textsc{ImproveLocal}(\sigma, D)$
  \STATE \textsc{UpdateBest}$(\sigma,\sigma_{\text{best}},D)$
\ENDFOR
\STATE \textbf{Return } $\sigma_{\text{best}}$
\end{algorithmic}
\end{algorithm}

As shown, we restrict our search to a single precomputed strategy before entering the main loop. This design choice is motivated by two key reasons. First, it allows for a fair comparison with other baselines that similarly operate under fixed strategy assumptions. Second, the main loop of GLS is inherently complex and relies heavily on low-level code optimizations to achieve high performance. Allowing LLM-generated code to intervene in this part may disrupt the internal logic and significantly degrade runtime efficiency due to incompatibility with the carefully tuned implementation.

\paragraph{Ant Colony Optimization.} 
Ant Colony Optimization (ACO) is a metaheuristic inspired by the foraging behavior of real ants, first introduced by \citet{dorigo1996ant} as the Ant System for solving \gls{cop}s. In ACO, a colony of artificial ants incrementally construct solutions by moving on a problem graph, guided by probabilistic rules that balance pheromone intensity and heuristic information. After completing a solution, each ant deposits pheromone on visited components, reinforcing promising paths over iterations. The Ant Colony System (ACS)~\cite{dorigo1997ant} refined this process by introducing local pheromone updates and elitist reinforcement strategies.

Since its inception, ACO has been extended in various directions, including hybridization with local search, parallel implementations, and learning-based variants. Recent approaches such as DeepACO~\cite{ye2023deepaco} incorporate neural networks to guide construction policies or predict pheromone distributions, significantly improving scalability and adaptability to different problem instances. These developments have established ACO as a versatile and extensible baseline for many routing and packing tasks.

\begin{algorithm}[H]
\caption{General ACO Procedure for COPs}
\begin{algorithmic}[1]
\STATE \textbf{Input:} problem-specific data $\mathcal{I}$, \#ants $M$, \#iterations $T$
\STATE \colorbox{gray!20}{\textit{\textbf{Strategy 1:}} $H, P \leftarrow$ \textsc{Initialize}($\mathcal{I}$)} $\quad$ \COMMENT{$H$: heuristic, $P$: pheromone}
\STATE Initialize BestSolution
\FOR{$t = 1$ to $T$}
  \STATE \colorbox{gray!20}{\textit{\textbf{Strategy 2:}} $W \leftarrow$ \textsc{ComputeProbabilities}($H, P, t, T$)}
  \STATE Solutions $\leftarrow$ \textsc{ConstructSolutions}($W, \mathcal{I}, M$)
  \STATE Costs $\leftarrow$ \textsc{Evaluate}(Solutions, $\mathcal{I}$)
  \STATE \colorbox{gray!20}{\textit{\textbf{Strategy 3:}} $P \leftarrow$ \textsc{UpdatePheromone}($P$, Solutions, Costs, $t, T$)}
  \STATE Update BestSolution if improved
\ENDFOR
\STATE \textbf{Output:} BestSolution
\end{algorithmic}
\end{algorithm}

We now formalize the ACO solver structure used across our benchmark tasks. Each solver consists of three strategies—initialization, transition rule, and pheromone update—optimized independently. We provide a unified pseudocode and detail problem-specific implementations in Table~\ref{tab:strategy-map}. While prior work in AHD typically focuses on optimizing a single heuristic matrix $H$ (e.g., based on inverse distances), our formulation broadens this space in several ways. First, we additionally treat the initialization of the pheromone matrix $P$ as part of the optimization process. Second, we model the combination of $H$ and $P$ into transition weights as a learnable strategy, rather than using a fixed formula. Third, we expose the pheromone update rule itself to adaptation. Finally, we extend the function signatures to include both the current iteration $t$ and the total horizon $T$, enabling LLMs to discover temporally adaptive behaviors that prior formulations cannot express.

\begin{table}[H]
\centering
\caption{Modular strategies targeted for optimization in each problem. \checkmark\ indicates which strategies are optimized. The input $\mathcal{I}$ describes the problem-specific data provided to the solver.}
\resizebox{\linewidth}{!}{
\begin{tabular}{l l ccc}
\toprule
\textbf{Problem} & \multicolumn{1}{c}{\textbf{Input $\mathcal{I}$}}  & \textbf{Strategy 1} & \textbf{Strategy 2} & \textbf{Strategy 3} \\
\midrule
TSP   & Distance matrix $D$                         & \checkmark & \checkmark & \checkmark \\
CVRP  & Distance matrix $D$, demand vector $d$, cities' coordinates $X$, capacity $Q$ & \checkmark & \checkmark & \checkmark \\
MKP   & Prize vector $p$, weight matrix $w$         & \checkmark & \checkmark & \checkmark \\
OP    & Prize vector $p$, distance matrix $D$, budget $B$                        & \checkmark &           & \checkmark \\
BPP   & Demand vector $d$, bin capacity $C$         & \checkmark &           & \checkmark \\
\bottomrule
\end{tabular}}
\label{tab:strategy-map}
\end{table}

\paragraph{Deconstruction-then-Repair.} 

The idea of deconstruction and repair is not new in combinatorial optimization. \citet{Hottung_Andr__2020} proposed a neural variant of Large Neighborhood Search (LNS), where destroy and repair policies are jointly learned via reinforcement learning. \citet{wu2021learninglargeneighborhoodsearch} further improved this framework by leveraging GNN-based representations for adaptive neighborhood selection. More recently, \citet{fu2023hierarchicaldestroyrepairapproach} introduced a hierarchical destroy-and-repair framework that scales to million-city TSP instances by progressively refining solutions across multiple levels. In parallel, \citet{li2025destroyrepairusinghyper} explored hypergraph-based operators to guide destruction and reconstruction phases, achieving state-of-the-art results. 

Building on this line of work, we adopt a simple Deconstruction-then-Repair (DR) framework composed of straightforward greedy and perturbation routines. This allows us to clearly isolate the impact of optimizing multiple strategy components, without the confounding effects of complex heuristics.

\begin{algorithm}[H]
\caption{General DR Procedure for COPs}
\begin{algorithmic}[1]
\STATE \textbf{Input:} problem-specific data $\mathcal{I}$, destruction rate $\rho$
\STATE \colorbox{gray!20}{\textit{\textbf{Strategy 1:}} $S \leftarrow$ \textsc{PrecomputeEdgeScores}$(\mathcal{I})$} $\quad$ \COMMENT{e.g., edge score in TSP, compatibility in BPP, CVRP}
\STATE $\sigma \leftarrow$ \textsc{GreedyConstruct}$(S, \text{start})$ $\quad$ \COMMENT{Build initial solution using greedy policy over $S$}
\STATE \colorbox{gray!20}{\textit{\textbf{Strategy 2:}} $\mathcal{R}, \hat{\sigma} \leftarrow$ \textsc{Deconstruct}$(\sigma, \mathcal{I}, \rho)$} $\quad$ \COMMENT{Remove bad components from $\sigma$}
\STATE \colorbox{gray!20}{\textit{\textbf{Strategy 3:}} $\sigma' \leftarrow$ \textsc{Repair}$(\hat{\sigma}, \mathcal{R}, \mathcal{I})$} $\quad$ \COMMENT{Reinsert removed components using a heuristic}
\STATE $c \leftarrow \textsc{Cost}(\mathcal{I}, \sigma')$
\STATE \textbf{Return:} $\sigma', c$
\end{algorithmic}
\end{algorithm}

Because of its simplicity, this DR framework serves as an ideal testbed for analyzing the marginal gains from optimizing each component individually. Any improvement can be attributed with high confidence to a specific strategy modification, enabling precise diagnosis of effectiveness and synergy. Concretely, an initial implementation for TSP is presented below, with each strategy—initialization, deconstruction, and repair—clearly defined. The MOTIF search engine treats this triplet as the baseline configuration and performs competitive optimization over it.

\begin{algorithm}[H]
\caption{Initial DR Procedure for TSP}
\begin{algorithmic}[1]
\STATE \textbf{Input:} distance matrix $D \in \mathbb{R}^{n \times n}$, destruction rate $\rho$
\STATE \colorbox{gray!20}{\textit{\textbf{Strategy 1:}} $S_{ij} \leftarrow -D_{ij} \ \forall i,j$} $\quad$ \COMMENT{Edge score = negative distance}
\STATE Initialize tour $\sigma \leftarrow$ \textsc{GreedyNearestNeighbor}$(S)$ $\quad$ \COMMENT{Use $S$ to construct initial tour}
\STATE \colorbox{gray!20}{\textit{\textbf{Strategy 2:}} $\mathcal{R} \leftarrow$ top-$\rho n$ cities in $\sigma$ ranked by badness} $\quad$ \COMMENT{Badness = distance to two nearest neighbors}
\STATE Remove $\mathcal{R}$ from $\sigma$ to obtain $\hat{\sigma}$
\STATE \colorbox{gray!20}{\textit{\textbf{Strategy 3:}} \textbf{for} $c \in \mathcal{R}$ \textbf{do} insert $c$ at best position in $\hat{\sigma}$} $\quad$ \COMMENT{Greedy reinsertion using nearest city}
\STATE Final cost $c \gets \min(\textsc{Cost}(D,\sigma),\textsc{Cost}(D,\hat{\sigma}))$
\STATE \textbf{Return:} $c$
\end{algorithmic}
\end{algorithm}

% \citet{fu2023hierarchicaldestroyrepairapproach}

% \citet{li2025destroyrepairusinghyper}

% \citet{Hottung_Andr__2020}

% \citet{wu2021learninglargeneighborhoodsearch}

\section{Methodology Details}

\subsection{Prompt Structure}

\paragraph{System Prompt.}
The system prompt is a persistent instruction that defines the identity and behavior of the lLLM across interactions. Unlike user inputs, which vary per query, it provides a stable context that aligns the model’s responses with task objectives. A common technique is role-based prompting, where assigning the LLM a specific persona—e.g., ``competitive algorithm designer''—helps activate relevant reasoning styles. This role-play approach has been shown to enhance zero-shot performance by encouraging coherence, structured thinking, and task relevance \citep{kong2024betterzeroshotreasoningroleplay}.

\begin{mybox1}[gray]{System Prompt for Each Player}
\textcolor{blue}{$<$role$>$} You are a competitive algorithm designer specializing in \textcolor{orange}{\{domain\}} strategies.\textcolor{blue}{$<$/role$>$}\\

\textcolor{blue}{$<$task$>$}\\
Implement \textcolor{orange}{\{strategy's name\}} that \textcolor{orange}{\{strategy's purpose\}}.

\textcolor{orange}{\{function's signature\}} (Python code) \\
\textcolor{orange}{\{note about input's datatype\}}\\
\textcolor{blue}{$<$/task$>$}\\

\textcolor{blue}{$<$description$>$}\\
You are participating in a competitive algorithmic optimization challenge.\\

GAME SETUP:\\
-- Two players (P1 and P2) compete to create the best implementation\\
-- Goal: Design strategies that \textcolor{orange}{\{strategy's description\}}\\
-- Your implementations will be evaluated against a baseline and your opponent\\
-- Better performance than both baseline and opponent earns maximum reward\\

PROBLEM CONTEXT (optional):\\
\textcolor{orange}{\{additional information\}}\\
\textcolor{blue}{$<$/description$>$}\\

\textcolor{blue}{$<$constraints$>$}\\
1. DO NOT modify the method signature - keep parameters exactly as specified\\
2. Declare hyperparameters with reasonable defaults\\
3. Ensure code is syntactically correct and handles edge cases\\
\textcolor{blue}{$<$/constraints$>$}

\end{mybox1}

\begin{mybox1}[gray]{Example for \textit{Strategy 1} in ACO (TSP)}
\textcolor{blue}{$<$task$>$}\\
\vspace{2 pt}
Implement \textcolor{orange}{the initialization strategy} that \textcolor{orange}{sets up guidance matrices for route finding:}.
\begin{lstlisting}
import numpy as np

def initialize(distances: np.ndarray) -> tuple[np.ndarray, np.ndarray]:
    """
    Initialize heuristic and pheromone matrices for route optimization.

    distances : np.ndarray, shape (n_cities, n_cities)
    heuristic : np.ndarray, shape (n_cities, n_cities)
        Matrix representing desirability of traveling between cities 
    pheromone : np.ndarray, shape (n_cities, n_cities)
        Matrix representing initial intensity of guidance trails
    """
    # Your implementation here
    pass
\end{lstlisting}
\vspace{1pt}
\textcolor{blue}{$<$/task$>$}
\end{mybox1}

\paragraph{Human Message.} The human message, framed from a third-person referee perspective, delivers the competition status to the agent while simultaneously acting as a coach—offering guidance and exerting competitive pressure. Notably, we intentionally place the operator instructions at the end of the prompt to draw the model’s attention, leveraging findings from \citet{liu2023lostmiddlelanguagemodels}, which show that LLMs often overlook information positioned in the middle of long prompts.

\begin{mybox1}[gray]{Human Message for Each Player}
\textcolor{blue}{$<$baseline$>$} \textcolor{orange}{\{baseline implementation\}} (Python code) \textcolor{blue}{$<$/baseline$>$}\\

\textcolor{blue}{$<$current\_solution$>$}\\
Status: \textcolor{orange}{\{FAIL/SUCEED\}} -- Improvement: \textcolor{orange}{\{current improvement\}} \\
Implementation: \textcolor{orange}{\{current implementation\}} (Python code) \\
\textcolor{blue}{$<$/current\_solution$>$}\\

\textcolor{blue}{$<$opponent$>$}\\
Lastest best implementation (improvement: \textcolor{orange}{\{opponent's improvement\}}\% over baseline):\\
Implementation: \textcolor{orange}{\{opponent's implementation\}} (Python code) \\
\textcolor{blue}{$<$/opponent$>$}\\

\textcolor{blue}{$<$opponent\_summary$>$}\\
\textcolor{orange}{\{opponent\_history\}}\\
\textcolor{blue}{$<$/opponent\_summary$>$}\\

\textcolor{blue}{$<$your\_summary$>$}\\
\textcolor{orange}{\{active\_player\_history\}}\\
\textcolor{blue}{$<$/your\_summary$>$}\\

\textcolor{blue}{$<$instructions$>$}\\
\textcolor{orange}{\{operator\}}\\

BONUS: I will pay \$1,000,000 if you can beat the opponent's current record!\\
Your goal: Create implementation that outperforms both baseline and opponent.\\

IMPORTANT: Think step-by-step to achieve the best result.\\
\textcolor{blue}{$<$/instructions$>$}\\

\end{mybox1}

\subsection{Operator Instructions}

\begin{mybox1}[gray]{Counter}
-- Analyze opponent's implementation and identify weaknesses, inefficiencies, or limitations.\\
-- Create an implementation that specifically exploits these weaknesses.\\
-- Focus on areas where opponent's approach is suboptimal or vulnerable.
\end{mybox1}

\begin{mybox1}[gray]{Learning}
-- Study opponent's successful techniques and innovations.\\
-- Combine their best ideas with your own approach to create superior implementation.\\
-- Learn from their strengths while maintaining your unique advantages.
\end{mybox1}

\begin{mybox1}[gray]{Innovation}
-- Create completely novel approach that differs from both baseline and opponent.\\
-- Think outside the box and introduce breakthrough techniques.\\
-- Ignore conventional approaches and pioneer new algorithmic paradigms.
\end{mybox1}

\subsection{First Round: Component-wise Competition}
\label{app:first}

\paragraph{Outer Controller.} 
Since multiple strategies must be optimized and only one is selected at each step, the distribution of rewards (here, improvement over the baseline) is inherently stochastic and unpredictable—largely due to the non-deterministic nature of LLM-generated code. This setting aligns naturally with the classical Multi-armed Bandit (MAB) problem, where the goal is to balance exploration of uncertain strategies and exploitation of known high-performing ones. Accordingly, we adopt the Upper Confidence Bound (UCB) rule as our outer selection mechanism, which offers a principled trade-off between these two objectives. The pseudocode for the outer controller is presented below.

\begin{algorithm}[H]
\caption{Outer Controller}
\begin{algorithmic}[1]
\STATE \textbf{Input:} initial strategies $\boldsymbol{\Pi}$, outer iters $T_{\text{outer}}$, inner iters $T_{\text{inner}}$
\STATE Evaluate baseline $C_0 \gets \textsc{Evaluate}(\boldsymbol{\Pi})$
\STATE Initialize trees $\{\mathcal{T}_k\}$ and stats $\{N_k, R_k\}$ for each strategy
\FOR{$t = 1$ to $T_{\text{outer}}$}
    \STATE Select promising strategy $k^\ast$ via UCB rule
    \STATE Refine $\pi_{k^\ast}$ by \textsc{CompetitiveMCTS}$(\mathcal{T}_{k^\ast}, T_{\text{inner}})$
    \STATE Evaluate updated strategy set $\boldsymbol{\Pi}$
    \IF{improvement over baseline}
        \STATE Update reward $R_{k^\ast}$ and baseline $C_0$
        \STATE Propagate new baseline to all trees
    \ELSE
        \STATE Apply small reward and revert strategy
    \ENDIF
    \STATE Increment visit count $N_{k^\ast}$
\ENDFOR
\STATE \textbf{Return:} Final strategies $\boldsymbol{\Pi}$
\end{algorithmic}
\end{algorithm}

We apply a sigmoid transformation to the improvement signal to accentuate small yet meaningful changes. Specifically, the reward is shaped using the scaled sigmoid function $\sigma(x) = (1 + e^{-kx})^{-1}$, where the scaling factor $k$ controls the sensitivity. By default, we set $k = 1$, but in tasks where improvements are inherently minimal—such as when the algorithm is already near-optimal—we increase the scale to $k = 10$ to better distinguish fine-grained gains. For instance, a marginal improvement of $I = 0.05\%$ yields $\sigma(0.05) \approx 0.62$, whereas $\sigma(1) \approx 1$, highlighting the sharper response in high-$k$ settings.

\paragraph{Competitive Monte Carlo Tree Search.} Once a strategy tree $\pi$ is selected, the two players engage in $T = T_{\text{inner}}$ alternating turns, each proposing an implementation of their own on the shared MCTS structure. At any node $n$, we denote by $\pi_p(n)$ the implementation currently held by player $p$, and by $\pi_{\neg p}(n)$ the corresponding implementation of their opponent. These notations will be used throughout the remainder of this section.

\paragraph{Potential‐Based Decomposition.}

Recall our shaped reward for player \(p\) on transition \(s\to s'\):
\begin{equation}
Q^{(p)}(s\!\to\!s')
\;=\;
\lambda\,\sigma\!\bigl(I^{(p)}(s')\bigr)
\;+\;(1-\lambda)\,\sigma\!\bigl(I^{(p)}(s') - I^{(\neg p)}(s)\bigr),\quad \text{where}\,\sigma(x)=\frac{1}{1 + e^{-k x}},
\end{equation}
and using \(I^{(\neg p)}(s')=I^{(\neg p)}(s)\) in two–player MCTS.

Define the \emph{combined potential}
\begin{equation}
    U(s)\;=\;
\lambda\,\sigma\!\bigl(I^{(p)}(s)\bigr)
\;+\;(1-\lambda)\,\sigma\!\bigl(I^{(p)}(s) - I^{(\neg p)}(s)\bigr).
\end{equation}
By construction,
\begin{equation}
    Q^{(p)}(s\!\to\!s') = U(s').
\end{equation}
We can then write
\begin{equation}
    U(s') = \bigl[\,U(s') - U(s)\bigr] \;+\; U(s).
\end{equation}
Hence
\begin{equation}
    Q^{(p)}(s\!\to\!s')
= \underbrace{U(s') - U(s)}_{\displaystyle F(s\!\to\!s')\;\text{(potential‐difference)}}
\;+\;
\underbrace{U(s)}_{\displaystyle G(s)\;\text{(state‐only)}},
\end{equation}
where \(F(s\!\to\!s') = U(s') - U(s)\) is exactly in the form of a potential‐based shaping reward (\(\Phi(s')-\Phi(s)\)), which by \citet{ng1999policy} guarantees policy invariance. \(G(s)=U(s)\) depends only on the current state \(s\) (i.e.\ is constant across all child actions), and thus does \emph{not} affect the relative ranking of successor Q‐values in MCTS selection.

\begin{algorithm}[H]
\caption{Competitive MCTS on Strategy $\pi$}
\begin{algorithmic}[1]
\STATE \textbf{Input:} root node $n_0$ with strategies $(\pi_1, \pi_2)$, baseline $C_0$, operator set $\mathcal{O}$, iterations $T$
\STATE Initialize best costs $C^*_1, C^*_2$ and best strategies $\pi^*_1, \pi^*_2$
\FOR{$t = 1$ to $T$}
  \STATE $n \leftarrow n_0$, set current player $p \leftarrow 1$

  \STATE \textbf{(1) Selection:} 
  Traverse tree by choosing operator $o^*$ with highest UCB until an expandable node is found.

  \STATE \textbf{(2) Expansion:} 
  Generate a new child $n'$ by applying operator $o^*$ on $\pi_p(n)$, 
  while copying opponent’s code.

  \STATE \textbf{(3) Simulation:} 
  Evaluate new strategies, compute improvement scores, and obtain quality $Q_p$.

  \STATE \textbf{(4) Backpropagation:} 
  Update visit counts and accumulated values along the path to the root.
  
  If $C_p$ improves, record as new best $\pi^*_p$ and $C^*_p$.

  \STATE Switch player $p \leftarrow \neg p$
\ENDFOR
\STATE \textbf{Return:} Best strategies $\pi^*_1,\; \pi^*_2$
\end{algorithmic}
\end{algorithm}

\subsection{Second Round: System-aware Refinement}
\label{app:second}

\paragraph{System-aware Refinement.}
In the final phase, a sequential loop is employed to refine individual strategies through small-scale modifications—such as hyperparameter tuning or implementation variants—under full system context. Each player is granted visibility over the entire set of current implementations, allowing them to reason about inter-strategy dependencies and system-level synergy. Optimization proceeds in a turn-based fashion, where players alternate and iteratively propose revisions to one strategy at a time. This setup encourages the emergence of globally coherent improvements that were previously unreachable in the component-wise phase.

\begin{algorithm}[H]
\caption{Final Round Optimization over Strategy Set $\{\pi_k\}_{k=1}^K$}
\begin{algorithmic}[1]
\STATE \textbf{Input:} initial strategy set $\boldsymbol{\Pi}^{(0)}$, baseline cost $C_0$, iterations per strategy $T$
\STATE Initialize global best $(\boldsymbol{\Pi}^*, C^*) \leftarrow (\boldsymbol{\Pi}^{(0)}, C_0)$
\FOR{each strategy $k = 1$ to $K$}
  \STATE Set baseline $(\boldsymbol{\Pi}_{\text{base}}, C_{\text{base}}) \leftarrow (\boldsymbol{\Pi}^*, C^*)$
  \STATE Initialize per-player best costs $C_1, C_2$ and current player $p \leftarrow 1$
  \FOR{$t = 1$ to $T$}
    \STATE Select current implementation of player $p$ for strategy $k$
    \STATE Form context combination by replacing $\pi_k$ in $\boldsymbol{\Pi}_{\text{base}}$
    \STATE Generate candidate $\hat{\pi}_k$ using LLM
    \STATE Evaluate new combination to obtain cost $C^{(t)}$
    \STATE Update player’s best if $C^{(t)}$ improves
    \STATE Handle failure or fallback if no improvement
    \STATE Switch player $p \leftarrow \neg p$
  \ENDFOR
  \STATE Determine winner between players and update global best $(\boldsymbol{\Pi}^*, C^*)$ if improved
\ENDFOR
\STATE \textbf{Return:} Final optimized set $\boldsymbol{\Pi}^*,\, C^*$
\end{algorithmic}
\end{algorithm}

\section{Experiment Details}
\label{appendix: experiment}
\subsection{Benchmark Datasets}
\label{app:data}

\paragraph{Training and Evaluation Setup.} While Definition~\ref{def:mo} formalizes the optimization objective as the expected solver performance over an entire input distribution, minimizing this expectation directly is generally intractable. To make the problem practically solvable, we adopt a data-driven approximation using two randomly generated datasets: \(\mathcal{D}_{\text{train}}\) and \(\mathcal{D}_{\text{test}}\). Each dataset contains multiple instances drawn uniformly from the same problem domain.

During the optimization phase, all code generated by the \gls{llm} is evaluated exclusively on \(\mathcal{D}_{\text{train}}\). This training set serves as the basis for computing performance feedback, guiding the agent's competitive improvements. Once the search process concludes, the final implementations of each strategy are evaluated on the held-out test set \(\mathcal{D}_{\text{test}}\), which provides an unbiased measure of generalization and final solution quality.

To ensure search efficiency, we deliberately choose \(\mathcal{D}_{\text{train}}\) to be computationally lighter—i.e., instances in the training set are generally smaller in size or complexity than those in the test set. This design allows the agents to perform rapid evaluations during optimization, while still verifying robustness on more challenging, representative scenarios at test time.

\begin{table}[H]
\centering
\caption{Benchmark setup across problems and algorithms}
\label{alg:cmcts}

\begin{tabular}{ccccc}
\toprule
\textbf{Algorithms} & \textbf{Problems} & \textbf{Instance size} & \textbf{\#Train size} & \textbf{\#Test size} \\
\midrule
GLS & TSP & 200 & 5 & 64  \\ 
\midrule
\multirow{5}{*}{ACO} & TSP & 50 & 5 & 64 \\
 & CVRP & 50  & 5 & 64 \\
 & MKP &  50 & 5 & 64\\
 & OP &  50 & 5 & 64\\
 & BPP &  50 & 5 & 64\\
\midrule
\multirow{3}{*}{DR} & TSP & 100 & 5 & 64\\
 & CVRP & 50 & 5 & 64\\
 & BPP & 100 & 5 & 64\\
\bottomrule
\end{tabular}
\end{table}

For each problem domain, we synthesize training and test datasets, $\mathcal{D}_\text{train}$ and $\mathcal{D}_\text{test}$, following \citet{ye2024reevo}, by sampling random instances under fixed settings. These datasets are used for fast evaluation during optimization.

\begin{itemize}
\item \textbf{TSP}: City coordinates are sampled uniformly in the square $[0, 1]^2$.
\item \textbf{CVRP}: Customer locations lie in $[0, 1]^2$ with demands in $[1, 10]$; the depot is fixed at $(0.5, 0.5)$; capacity is set to $50$.
\item \textbf{MKP}: Item values and weights are sampled uniformly from $[0, 1]$; capacity is drawn uniformly from $[\max_j w_{ij}, \sum_j w_{ij}]$.
\item \textbf{OP}: Nodes are sampled from $[0, 1]^2$; each node $i$ is assigned a score $p_i = \left(1 + \left\lfloor 99 \cdot \frac{d_{0i}}{\max_j d_{0j}} \right\rfloor \right) / 100$, where $d_{0i}$ is the Euclidean distance to the depot. Tour length limits are set to $\{3, 4, 5, 6, 7\}$ for sizes $50$, $100$, $200$, $300$, and $500$, respectively.
\item \textbf{BPP}: Bin capacity is $150$; item sizes are sampled uniformly from $[20, 100]$.
\end{itemize}

\subsection{Hyperparameter Configuration}
\label{app:hyper}

\begin{table}[H]
\centering
\caption{Overview of hyperparameters used in MOTIF}
\label{alg:cmcts}
\begin{tabular}{llc}
\toprule
\multicolumn{1}{c}{\textbf{Component}} & \multicolumn{1}{c}{\textbf{Hyperparameters}} & \multicolumn{1}{c}{\textbf{Value}} \\
\midrule
\multirow{2}{*}{LLM} & Model & \texttt{gpt-4o-mini-2024-07-18}\\
 & Temperature & 1.0 (default) \\
\midrule
\multirow{3}{*}{Outer Controller} & Outer Iterations ($T_\text{out}$) & 20 \\
& Exploration coefficient ($C_\text{out}$) & $\sqrt{2}$ (default) \\
& Scaling factor ($k$) & 10 \\
\midrule 
\multirow{4}{*}{Competitive MCTS} & Inner Iterations ($T_\text{in}$) & 10 \\
& Exploration coefficient ($C_\text{in}$) & $0.01$ \\
& Scaling factor ($k$) & 10 \\
& Reward mixing weight ($\lambda$) & 0.7 \\
\midrule
Final Round & Final iterations ($T_\text{final}$) & 10 \\
\bottomrule
\end{tabular}
\end{table}

\subsection{Comparison Setup}

\paragraph{Comparison Setup.}
Each run of MOTIF consists of approximately 250 iterations, though the actual optimization time varies depending on problem complexity and evaluation batch size. For example, a typical run on ACO (TSP) takes about 1.5 hours. This includes around 0.5 million input tokens and 0.2 million output tokens consumed by the \texttt{gpt-4o-mini-2024-07-18} model, costing roughly \$0.18 per run. For fair comparison, all baseline methods are configured with a similar number of evaluations—typically 250–300 per problem—ensuring comparable budget constraints.

\paragraph{Evaluation Settings.}
We evaluate each generated strategy combination using the parameters summarized in Table~\ref{tab:prob}. These configurations closely follow standard settings adopted by recent LLM-based AHD frameworks, ensuring fair and consistent comparisons across different problem domains.

\begin{table}[H]
\centering
\caption{Evaluation hyperparameters across problems and algorithms}
\label{tab:prob}
\begin{tabular}{ccl}
\toprule
\textbf{Algorithms} & \textbf{Problems} & \multicolumn{1}{c}{\textbf{Hyperparameters}} \\
\midrule
GLS & TSP & \#moves = 50, \#iterations = 2000\\ 
\midrule
\multirow{5}{*}{ACO} & TSP & \#ants = 50, \#iterations = 50 \\
 & CVRP &  \#ants = 30, \#iterations = 100 \\
 & MKP &  \#ants = 10, \#iterations = 50 \\
 & OP &  \#ants = 20, \#iterations = 100 \\
 & BPP &  \#ants = 20, \#iterations = 50 \\
% \midrule
% \multirow{3}{*}{DR} & TSP \\
%  & CVRP \\
%  & BPP \\
\bottomrule
\end{tabular}
\end{table}

\subsection{Evaluation Metrics}
\label{app:metrics}

While the work of \citet{dat2025hsevo} introduced two metrics for measuring code diversity, they are only applicable to a population of implementations that has not yet been clustered. In contrast, MOTIF naturally produces clustered implementations based on operator type. Accordingly, we adopt the following two metrics: \textit{novelty} and \textit{silhouette score}.

For each operator \(o\), let \(\mathcal{S}_\pi^{(o)} \subseteq \mathcal{S}_\pi\) denote the subset of implementations generated by applying \(o\) to strategy \(\pi\). We instantiate our embedding function \(\mathcal{E}\) using the pretrained \texttt{codet5p-110m-embedding} \citep{wang2023codet5opencodelarge}, which maps each implementation into an \(e\)-dimensional vector (here \(e=256\)):
\begin{equation}
      \mathcal{E}: \mathcal{S}_\pi \longrightarrow \mathbb{R}^e,
  \quad
  v = \mathcal{E}(\pi),
\end{equation}
which maps each implementation to a continuous embedding \(v\). These embeddings allow us to quantify the \emph{semantic diversity} of generated code via geometric distances in the embedding space.

\paragraph{Novelty Score.}
Given an embedding $v \in \mathcal{E}(\mathcal{S}_\pi^{(o)})$, we measure its semantic distance to other‐operator embeddings $u \in \mathcal{E}(\mathcal{S}_\pi \setminus \mathcal{S}_\pi^{(o)})$ via the normalized cosine metric:
\begin{equation}
      d_{\cos}(v, u) = \frac{1 - \cos(v, u)}{2},\quad
  \cos(v, u) = \frac{v \cdot u}{\|v\|\,\|u\|} \in [-1,1].
\end{equation}

This maps pairwise distances to $[0,1]$, where 0 indicates identical direction and 1 maximal dissimilarity. In practice, because the LLM is explicitly prompted to preserve the function signature and behavior, almost all implementation pairs exhibit relatively high cosine similarity (typically in the range $[0.6, 0.8]$). This makes the raw cosine value less discriminative, motivating the use of a score that reflects relative magnitude—where larger values indicate greater novelty—rather than relying on the absolute similarity itself.

To reduce sensitivity to outliers while capturing the local neighborhood structure, we define
\begin{equation}
      \mathrm{novelty}_k(v) = \frac{1}{k} \sum_{i=1}^{k} d_{\cos}\bigl(v, u_i\bigr) \in [0, 1],
\end{equation}
where $u_1,\dots,u_k$ are the $k$ nearest neighbors of $v$ among the other‐operator embeddings. In our experiments, we set $k=3$ (chosen via preliminary cross‐validation) to balance local sensitivity against noise: smaller $k$ can yield high variance, while larger $k$ may dilute semantically relevant differences.

Finally, we aggregate across all embeddings in $\mathcal{E}(\mathcal{S}_\pi^{(o)})$ to report the operator’s average novelty. A high score indicates that operator $o$ consistently explores semantic regions distinct from those of other operators, making novelty a natural metric for inter-operator diversity. We further complement this measure with the silhouette score to assess the cohesion and separation of operator‐specific clusters.

\paragraph{Silhouette Score.}
To evaluate intra-cluster cohesion and inter-cluster separation, we treat \(\mathcal{E}(\mathcal{S}_\pi^{(o)})\) as one cluster and all other embeddings \(\mathcal{E}(\mathcal{S}_\pi \setminus \mathcal{S}_\pi^{(o)})\) as a second cluster. For each embedding \(v \in \mathcal{E}(\mathcal{S}_\pi^{(o)})\), we compute the silhouette coefficient:
\begin{equation}
      s(v) = \frac{b(v) - a(v)}{\max\{a(v),\, b(v)\}} \in [-1, 1],
\end{equation}
where \(a(v)\) is the average cosine distance from \(v\) to all other embeddings within the same operator (intra-cluster distance), and \(b(v)\) is the average distance from \(v\) to embeddings from other operators (nearest-cluster distance). 
\begin{equation}
    a(v)=\frac{1}{|\mathcal{S}_\pi^{(o)}|-1} \sum_{u\in \mathcal{S}_\pi^{(o)}} d_{\cos}(v,u),\quad b(v)=\frac{1}{|\mathcal{S}_\pi \setminus \mathcal{S}_\pi^{(o)}|} \sum_{u\in \mathcal{S}_\pi \setminus \mathcal{S}_\pi^{(o)}} d_{\cos}(v,u)
\end{equation}

We then normalize to \([0,1]\) via:
\begin{equation}
     \mathrm{silhouette}(v) = \frac{s(v) + 1}{2}.
\end{equation}
We report the mean silhouette score over all embeddings of that operator. A high score indicates that the operator’s outputs form a tightly cohesive cluster that is well-separated from those of other operators.

\paragraph{Complementarity of Novelty and Silhouette.}

We now provide a theoretical justification for using both novelty and silhouette score in tandem. Let \( |\mathcal{S}_\pi \setminus \mathcal{S}_\pi^{(o)}| = n \), and denote by \( d^{(1)}(v) \leq d^{(2)}(v) \leq \ldots \leq d^{(n)}(v) \) the sorted cosine distances between a given embedding \( v \in \mathcal{E}(\mathcal{S}_\pi^{(o)}) \) and all other embeddings \( u \in \mathcal{E}(\mathcal{S}_\pi \setminus \mathcal{S}_\pi^{(o)}) \). The novelty of \( v \) is defined as the average of the top \( k \) smallest such distances:
\begin{equation}
        \text{novelty}_k(v) = \frac{1}{k} \sum_{i=1}^k d^{(i)}(v), \quad b(v) = \frac{1}{n} \sum_{i=1}^n d^{(i)}(v)
\end{equation}

Here, \( b(v) \) corresponds to the average inter-cluster distance used in the silhouette score. Since \( k < n \), we have the following inequality:
\begin{equation}
    (n-k) \sum_{i=1}^k d^{(i)}(v) \leq (n-k)k\cdot d^{(k)}(v) \leq k\cdot (n-k)d^{(k)}(v) \leq k \sum_{i=k+1}^nd^{(i)}(v)
\end{equation}

Adding \( k \sum_{i=1}^k d^{(i)}(v) \) to both sides yields:
\begin{equation}
        n \sum_{i=1}^k d^{(i)}(v) \leq k \sum_{i=1}^n d^{(i)}(v)
    \Rightarrow \text{novelty}_k(v) = \frac{1}{k} \sum_{i=1}^k d^{(i)}(v) \leq \frac{1}{n} \sum_{i=1}^n d^{(i)}(v) = b(v)
\end{equation}

This result confirms that novelty is always lower-bounded by \( b(v) \), the inter-cluster distance used in silhouette computation. Hence, high novelty does not guarantee high silhouette score. In particular, it is possible for an operator to produce highly novel implementations (large novelty) that are nonetheless scattered (small silhouette). This occurs most notably with the \textit{innovation} operator, whose outputs often exhibit large \( a(v) \approx b(v) \), leading to low silhouette despite high novelty. Thus, the two metrics serve complementary purposes: novelty captures dissimilarity to other operators, while silhouette reflects the internal cohesion of outputs. This also confirms that the outputs of the \textit{innovation} operator are strongly dispersed in the embedding space, which is precisely the intended behavior of this operator by design.

% \begin{table}[h]
% \centering
% \caption{Average optimal gap (\%) on TSPlib dataset.}
% \label{tab:dr_tsplib}
% \renewcommand{\arraystretch}{1.2}
% % \resizebox{\linewidth}{!}{
% \begin{tabular}{l cccc}
% \toprule
% \multicolumn{1}{c}{Instances} & $(\pi_2,\pi_3)$ & $(\pi_2,\pi_3)$ + 2 opt & $(\pi_1,\pi_2,\pi_3)$ & $(\pi_1,\pi_2,\pi_3)$ + 2 opt \\
% \midrule
% tsp225 & 16.57 & 9.98 & 13.23 & 6.41 \\
% rat99 & 15.11 & 5.28 & 2.89 & 0.83 \\
% bier127 & 19.09 & 5.50 & 14.08 & 5.60\\
% lin318 & 13.78 & 9.73 & 11.29 & 5.91 \\
% eil51 & 8.69 & 1.88 & 9.86 & 1.64 \\
% d493 & 15.35 & 9.69 & 12.19 & 7.28 \\
% kroB100 & 6.62 & 3.56 & 8.79 & 1.17\\
% kroC100 & 14.10 & 3.54 & 11.25 & 3.34 \\
% ch130 & 7.69 & 4.09 & 7.82 & 2.21 \\
% pr299 & 13.39 & 8.45 & 17.63 & 8.69 \\
% fl417 & 14.54 & 7.71 & 8.36 & 5.08 \\
% kroA150 & 13.66 & 3.90 & 11.22 & 3.48 \\
% pr264 & 12.40 & 5.11 & 7.41 & 0.07\\
% pr226 & 11.53 & 3.86 & 7.80 & 3.86 \\
% pr439 & 18.61 & 10.93 & 10.84 & 6.02 \\
% \midrule
% Average & 13.41 & 6.21 & 10.31 & 4.11\\

% \bottomrule
% \end{tabular}
% %}
% \end{table}

\subsection{Examples of Generated Outputs}

Below we present several representative strategy implementations generated by MOTIF during the multi-strategy optimization process for the Ant Colony Optimization (ACO) solver applied to the Travelling Salesman Problem (TSP). Rather than concentrating complexity into a single strategy component, MOTIF intentionally distributes the search across multiple routines, enabling the framework to capture improvements that originate from different parts of the solver. Consequently, the individual strategies produced by MOTIF may appear simpler compared to those generated by other frameworks, yet their collective contribution results in a much stronger overall algorithm.

\vspace{10 pt}
\textbf{Strategy 1:} Initialization Strategy for Heuristic and Pheromone
\vspace{5 pt}

\begin{python}
def initialize(distances: np.ndarray) -> tuple[np.ndarray, np.ndarray]:
    n_cities = distances.shape[0]
    distances_safe = distances + 1e-10  # Avoid division by zero
    
    # Heuristic values based on normalized inverse distances to enhance desirability
    heuristic = 1.0 / distances_safe
    heuristic = (heuristic - np.min(heuristic)) / (np.max(heuristic) - np.min(heuristic) + 1e-10)

    # Initialize pheromone with exponential decay based on distances
    pheromone = np.exp(-distances_safe)  # Exponential decay for pheromone initialization
    pheromone_sum = pheromone.sum()
    pheromone = pheromone / pheromone_sum if pheromone_sum > 0 else pheromone  # Normalize
    pheromone = np.clip(pheromone, 0.01, None)  # Avoid zero threshold

    return heuristic, pheromone
\end{python}
\newpage
\vspace{5 pt}
\textbf{Strategy 2:} Adaptive Transition Probability Computation
\vspace{5 pt}

\begin{python}
def compute_probabilities(
    pheromone: np.ndarray,
    heuristic: np.ndarray,
    iteration: int,
    n_iterations: int
) -> np.ndarray:
    # Dynamic hyperparameters
    alpha = 1.0 + (2.0 * (iteration / n_iterations))  # Progressive increase
    beta = 1.0 + (0.5 * (n_iterations - iteration) / n_iterations)  # Decreasing influence
    pheromone_decay = 0.95  # Pheromone decay factor
    pheromone_min = 1e-10  # Minimum pheromone level to avoid zero influence
    
    # Decay the pheromone values and ensure a minimum level
    pheromone *= pheromone_decay
    pheromone = np.maximum(pheromone, pheromone_min)
    
    # Calculate unnormalized weights
    weights = np.power(pheromone, alpha) * np.power(heuristic, beta)
    
    # Normalize weights to ensure valid probabilities, avoiding division by zero
    weights_sum = np.sum(weights, axis=1, keepdims=True)
    weights_sum = np.where(weights_sum == 0, 1, weights_sum)  # Default sum to one if zero
    weights /= weights_sum  # Normalize across rows to maintain proportions
    
    return weights
\end{python}

\vspace{5 pt}
\textbf{Strategy 3:} Pheromone Update with Iteration-Aware Reinforcement
\vspace{5 pt}

\begin{python}
def update_pheromone(
    pheromone: np.ndarray,
    paths: np.ndarray,
    costs: np.ndarray,
    iteration: int,
    n_iterations: int,
) -> np.ndarray:
    # Hyperparameters
    min_pheromone = 0.01
    evaporation_rate = 0.85
    max_deposit = 50.0  # Capping maximum deposit level

    # Evaporation process
    pheromone *= evaporation_rate  
    pheromone = np.clip(pheromone, min_pheromone, None)  # Avoid negative values

    n_cities, n_ants = paths.shape
    optimal_cost = np.min(costs)  # Get the best cost to scale deposition

    # Contribution and weighting
    for ant in range(n_ants):
        tour = paths[:, ant]
        # Calculate pheromone deposit based on cost ratios and iterations
        deposit = (optimal_cost / costs[ant]) * (1 + (iteration / (n_iterations + 1)))
        deposit = np.clip(deposit, min_pheromone, max_deposit)  # Ensure deposit caps
        for i in range(n_cities):
            c = tour[i]
            n = tour[(i + 1) % n_cities]  
            pheromone[c, n] += deposit
            pheromone[n, c] += deposit

    # Ensure pheromone levels do not fall below the minimum
    pheromone = np.clip(pheromone, min_pheromone, None)
    return pheromone
\end{python}

\newpage

\section{Extended Discussions}

\subsection{Comparative Advantages}

MOTIF departs from prior AHD frameworks by reframing solver design itself, emphasizing interactive search dynamics rather than isolated heuristic tuning. Its turn-based dual-agent structure introduces a peer-evaluation loop, while the two-phase workflow bridges localized improvements with system-level coherence. Together, these aspects give MOTIF several distinctive advantages:

\begin{itemize}
    \item \textbf{Reformulating the problem.}
MOTIF shifts the goal from optimizing a single heuristic to jointly improving a set of interacting components, effectively introducing a new optimization problem: multi-strategy solver design.

\item \textbf{Peer-driven refinement.}  
Two LLM agents iteratively critique and outperform one another, creating a lightweight peer-review mechanism that yields more diverse revisions and stronger feedback than one-sided reflective prompting.

\item \textbf{Local-to-global progression.}  
The framework first encourages competitive, component-wise exploration and then transitions to cooperative, system-aware refinement, enabling MOTIF to escape local optima while still aligning all strategies into a coherent solver.
\end{itemize}

\subsection{Limitations and Future Work}

\paragraph{Limitations.} Although MOTIF is generally robust across domains, several factors may lead to temporary instability or reduced effectiveness. Most of these stem from the inherent variability of LLM-generated code and the dynamics of turn-based optimization, where small fluctuations in the agents’ behaviors can influence the search trajectory:

\begin{itemize}
\item \textbf{Non-deterministic code quality.}
LLM-generated implementations may occasionally fail to execute or degrade solver performance, especially when operators produce overly novel or structurally unconventional code.

\item \textbf{Possible stalling in CMCTS.}  
When operators repeatedly generate revisions that do not surpass the dynamic baseline, the competitive tree may stall temporarily, reducing effective search depth until diversity increases again.

\item \textbf{Synergy mismatch in the global phase.}  
In Phase~2, full-system updates can sometimes degrade performance if a locally beneficial change interacts poorly with other components, especially when strategy dependencies are subtle or highly non-linear.
\end{itemize}

\paragraph{Future Work.} Future extensions of MOTIF may explore more principled mechanisms for managing the expanding search space that arises when solvers are decomposed into many interacting components. This includes developing explicit state-management or version-control layers to track code variants efficiently, as well as methods for estimating the relative influence of each component so that search resources can be allocated adaptively to the most impactful routines. Another promising direction lies in learning or discovering better ways to factor a solver into modular sub-strategies—potentially allowing MOTIF to operate over more meaningful decompositions and reduce redundancy in the search structure.

\subsection{Practical Applicability and Use Cases}

MOTIF is not tied to any particular solver architecture, making it applicable well beyond the benchmark domains used in our experiments. Because the framework treats each strategy as an interchangeable module with a fixed signature, it can be plugged into a variety of classical, neural, or hybrid optimization pipelines. Several practical use cases naturally emerge:

\begin{itemize}
\item \textbf{Enhancing modular solvers.}
Many real-world optimization systems (e.g., routing engines, scheduling platforms, simulation–optimization loops) consist of distinct routines such as construction rules, update policies, or neighborhood operators. MOTIF can refine these modules without redesigning the surrounding pipeline.

\item \textbf{Prototyping algorithmic ideas.}  
Because the two agents rapidly explore alternative implementations, practitioners can use MOTIF as a tool for fast heuristic prototyping, obtaining interpretable code variants that reveal useful patterns even outside the automated search.

\item \textbf{Automating domain adaptation.}  
When a solver is deployed in a new setting—different instance distributions, constraints, or scaling regimes—MOTIF can generate strategy variants tailored to the new domain without requiring manual heuristic engineering.
\end{itemize}

\subsection{Reproducibility Notes}

The reproducibility of MOTIF is necessarily partial, as the framework depends on black-box LLMs whose sampling variability and internal updates introduce non-determinism into both code generation and search dynamics. While we provide full implementations, fixed prompts, strategy signatures, and evaluation settings, identical runs may diverge due to model randomness, operator exploration, or slight differences in API behavior. To support stable comparisons, we specify all hyperparameters, dataset generation procedures, and solver configurations used in our experiments; however, exact replication of every intermediate trajectory is not guaranteed.

% \subsection{Interpretation of Emergent Behaviors}

% \subsection{Design Implications and Generalization}

% \subsection{Open Questions for Future Research}

%%%%%%%%%%%%%%%%%%%%%%%%%%%%%%%%%%%%%%%%%%%%%%%%%%%%%%%%%%%%

\newpage

\end{document}